\documentclass[conference]{IEEEtran}
\usepackage{cite}
\usepackage{amsmath,amssymb,amsfonts}
\usepackage{algorithmic}
\usepackage{graphicx}
\usepackage{textcomp}
\usepackage{xcolor}
\usepackage{booktabs}
\usepackage{rotating}
\usepackage{makecell}
\usepackage{tabularray}
\usepackage{adjustbox}
\UseTblrLibrary{booktabs}
\usepackage{placeins}
\usepackage{bm}
\def\BibTeX{{\rm B\kern-.05em{\sc i\kern-.025em b}\kern-.08em
    T\kern-.1667em\lower.7ex\hbox{E}\kern-.125emX}}

\DeclareMathOperator*{\argmin}{arg\,min}
\usepackage[colorlinks = true,
            linkcolor = blue,
            urlcolor  = blue,
            citecolor = black,
            anchorcolor = black]{hyperref}

\newcommand\blfootnote[1]{%
  \begingroup
  \renewcommand\thefootnote{}\footnote{#1}%
  \addtocounter{footnote}{-1}%
  \endgroup
}

\begin{document}

\title{SoK: A Review of Differentially Private Linear Models For High-Dimensional Data}

\author{\IEEEauthorblockN{Amol Khanna}
\IEEEauthorblockA{\textit{Booz Allen Hamilton} \\
Annapolis Junction, MD, USA \\
\texttt{Khanna\_Amol@bah.com}}
\and
\IEEEauthorblockN{Edward Raff}
\IEEEauthorblockA{\textit{Booz Allen Hamilton} \\
\textit{University of Maryland, Baltimore County}\\
Annapolis Junction, MD, USA \\
\texttt{Raff\_Edward@bah.com}}
\and 
\IEEEauthorblockN{Nathan Inkawhich}
\IEEEauthorblockA{\textit{Air Force Research Laboratory} \\
Rome, NY, USA \\
\texttt{Nathan.Inkawhich@us.af.mil}}
}

\maketitle

\begin{abstract}
Linear models are ubiquitous in data science, but are particularly prone to overfitting and data memorization in high dimensions. To guarantee the privacy of training data, differential privacy can be used. Many papers have proposed optimization techniques for high-dimensional differentially private linear models, but a systematic comparison between these methods does not exist. We close this gap by providing a comprehensive review of optimization methods for private high-dimensional linear models. Empirical tests on all methods demonstrate robust and coordinate-optimized algorithms perform best, which can inform future research. Code for implementing all methods is released online. 
\end{abstract}


\section{Introduction}

\blfootnote{Approved for Public Release; Distribution Unlimited. PA Number: AFRL-2023-5408}Linear models, like linear and logistic regression, are ubiquitous in current data science and data analytics efforts. Their simple structure enables them to train quickly and generalize well on simple problems. Additionally, they can be interpreted easily to understand model decision making, which is essential in regulated fields like medicine and finance. For example, linear regression has been used to predict future consumer demand, corporate resource requirements, and house prices \cite{lyssiotou2002age,madhuri2019house}. Logistic regression has been used for disease prediction and fraud detection \cite{kurt2008comparing,sahin2011detecting}. 

In the low-dimensional regime, where a dataset has many more datapoints than features, linear models typically generalize well without significant tuning. From an information-theoretic perspective, this is because the model has enough data to learn major trends in the dataset, which should generalize to future instances. This is why linear models are often most effective on simple, large datasets \cite{murphy2022probabilistic}. 

However, many modern datasets are high-dimensional, having more features than datapoints. This is common in genomics or finance, where the expression of many genes or prices of many assets outnumber individual observations. In this case, linear models can overfit to the data, producing poor generalization to future inputs \cite{buhlmann2013statistical}. 

A common solution to this problem uses regularization, which constrains the weight vector of these models. For example, a common goal is to constrain the number of nonzero coefficients of the model's weight. This both promotes generalization and produces a weight vector which highlights the features which are most strongly related to the model's output \cite{rigollet201518}. 

However, linear models' solutions can produce weights which rely on information contained in one or a few datapoints. This can be problematic for models trained on sensitive data, such as those used in medicine or banking, since the parameters of these models can leak information about these datapoints. Specifically, extensive research has demonstrated that membership inference attacks can reconstruct datapoints used in models with high accuracy given access to only the model's outputs or parameters. This is especially true in the high-dimensional regime, where models are easily capable of overfitting to data \cite{hu2022high}. 

A theoretical solution to this problem employs differential privacy. Differential privacy is a statistical guarantee of the privacy of an algorithm's outputs, which ensures that an algorithm does not rely too heavily on information in any one datapoint. Differential privacy prevents membership inference and dataset reconstruction attacks on a trained model \cite{dwork2006differential}. 

Deferentially private linear and logistic regression have been studied extensively in the past two decades, but primarily from a theoretical perspective. Many optimization strategies and heuristics have been developed, but for most important statistical tasks, systematic comparisons between different algorithms have not been conducted. 

This paper reviews methods to develop differentially private high-dimensional linear models. This task is fundamental in private statistics since linear models are typically the first models tested in a data science pipeline. Since differential privacy works best with simple models, using an appropriately trained linear model can avoid false conclusions on linear models' ineffectiveness for certain problems. Finally, differential privacy typically struggles with high-dimensional problems since noise can overwhelm the signals in these problems. This review seeks to identify whether specific optimization methods consistently improve performance on private high-dimensional problems. 

Specifically, this review contributes the following: 
\begin{enumerate}
    \item We provide the first centralized review of all methods performing high-dimensional linear modeling with differential privacy, and provide insights about methods' strengths, weaknesses, and assumptions. 
    \item We implement methods reviewed in code and release all code at \href{https://github.com/afrl-ri/differential-privacy-review}{https://github.com/afrl-ri/differential-privacy-review}. Many reviewed works do not release code, which can make it challenging for their methods to be tested and improved upon by other works. We close this gap. 
    \item We provide a systematic empirical comparison of each method's performance on a variety of datasets. Current differential privacy literature typically compares methods' theoretical utility. While this is a useful metric, empirical performance is often overlooked. In conducting these empirical tests, we find surprising and previously unidentified trends in performance which can influence future research. 
\end{enumerate}

Finally, we remark that in this article, the term ``linear models'' refers to both linear regression, where the output is an unconstrained numeric prediction (i.e., the mean-squared error loss), and logistic regression, where a binary label is desired. Most proposed methods in the literature support both, but some linear regression and logistic regression specific methods will also be evaluated. 

\section{Preliminaries}

In this section we provide brief overviews of differential privacy and nonprivate optimization methods for high-dimensional linear models. This provides context for our literature review in section III. 

\subsection{Differential Privacy}

Differential privacy (DP) is a statistical guarantee of the privacy of an algorithm's outputs. Specifically, given an algorithm $\mathcal{A}$, it is $(\epsilon, 0)$ DP if for all datasets $\mathcal{D}$ and $\mathcal{D}'$ differing on one datapoint and any event $E$, $$\frac{\mathbb{P}[\mathcal{A}(\mathcal{D}) \in E]}{\mathbb{P}[\mathcal{A}(\mathcal{D}') \in E]} \leq e^{\epsilon}.$$
In other words, DP bounds the amount $\mathcal{A}$'s output can depend on a single datapoint. Since it cannot rely too much on any individual datapoint, it cannot reveal significant information about a single individual in a dataset \cite{dwork2006differential}. 

Implementing the above definition of DP for a nonprivate algorithm $\mathcal{B}$ requires bounding its $L_1$ sensitivity. This is the maximum amount that $\mathcal{B}$'s output can change between two neighboring datasets as measured by the $L_1$ norm. Mathematically, this is written as $$\max_{\mathcal{D}, \mathcal{D}': \ \text{dist}(\mathcal{D}, \mathcal{D}') = 1} \lVert \mathcal{B}(\mathcal{D}) - \mathcal{B}(\mathcal{D'}) \rVert_1.$$ Once the sensitivity is known, Laplacian noise with appropriate scale can be added to $\mathcal{B}$ to produce a DP version of $\mathcal{B}$ \cite{dwork2014algorithmic}. However, adding Laplacian noise to $\mathcal{B}$ using its $L_1$ sensitivity can destroy its efficacy. The noise can corrupt its outputs so much that it becomes useless. To address this issue, approximate DP was developed. 

Approximate differential privacy, otherwise known as $(\epsilon, \delta)$ DP, modifies the above definition of DP to $$\mathbb{P}[\mathcal{A}(\mathcal{D}) \in E] \leq e^{\epsilon}\mathbb{P}[\mathcal{A}(\mathcal{D}') \in E] + \delta,$$ where $0 < \delta \leq 1$. In this case, a nonprivate algorithm $\mathcal{B}$ can be made $(\epsilon, \delta)$ DP by adding noise from a normal distribution with scale proportional to $\mathcal{B}$'s $L_2$ sensitivity. This method can add significantly less noise to $\mathcal{B}$'s outputs because the Gaussian distribution's tails are significantly lighter than the Laplace distribution \cite{beimel2013private}. 

Finally, we mention some important properties of DP employed in the methods reviewed in section III. We begin with the post-processing property, which states that if $\mathcal{A}(\mathcal{D})$ is DP, then for any function $g$ which does not re-access the dataset $\mathcal{D}$, $g(\mathcal{A}(\mathcal{D}))$ is DP with the same privacy parameters \cite{dwork2014algorithmic}. 

Next, we discuss composition. Say $\mathcal{A}(\mathcal{D})$ is $(\epsilon, \delta)$ DP, where $0 \leq \delta \leq 1$. Then under parallel composition, where $\mathcal{D}$ is broken into disjoint sets such that $\mathcal{D}_1 \cup \mathcal{D}_2 \cup \dots \cup \mathcal{D}_k = \mathcal{D}$, releasing all outputs $\mathcal{A}(\mathcal{D}_1), \dots \mathcal{A}(\mathcal{D}_k)$ is DP with the same parameters as $\mathcal{A}(\mathcal{D})$ \cite{dwork2009differential}. In contrast, under sequential composition, if $\mathcal{A}_1(\mathcal{D})$ is $(\epsilon_1, \delta_1)$ DP and $\mathcal{A}_2(\mathcal{D})$ is $(\epsilon_2, \delta_2)$ DP, then releasing $\mathcal{A}_1(\mathcal{D})$ and $\mathcal{A}_2(\mathcal{D})$ is $(\epsilon_1 + \epsilon_2, \delta_1 + \delta_2)$ DP \cite{kairouz2015composition}. 

Note that the above bound for sequential composition is called the basic composition theorem. Employing concentration inequalities, using a small value $\delta'$, the advanced composition theorem states that we can produce a sublinear increase in $\epsilon$. For example, for $k$-fold adaptive composition under $(\epsilon, 0)$ DP, the advanced composition theorem states that the output is $(\epsilon', \delta')$ DP, where $\epsilon' = 2\epsilon \sqrt{2k \log(1/\delta')}$. Similarly, for adaptive composition of $(\epsilon, \delta)$ DP algorithms, the advanced composition theorem shows a $k$-fold composition is $(\epsilon', k\delta + \delta')$ DP \cite{kairouz2015composition}. 

The advanced composition theorem is tight for $(\epsilon, 0)$ DP, meaning that it produces the lowest privacy parameters which can generally be applied to any $(\epsilon, 0)$ DP mechanism. However, advanced composition is not tight for $(\epsilon, \delta)$ DP where $\delta > 0$ \cite{near2021programming}. This has led to the adoption of R\'enyi and zero concentrated DP, reformulations of DP which are naturally able to produce tight Gaussian mechanism composition for $(\epsilon, \delta)$ mechanisms \cite{bun2016concentrated,mironov2017renyi}. 

Finally, many DP mechanisms have been developed, but here we mention one specifically: the report-noisy-max  mechanism. The report-noisy-max mechanism is one application of the exponential mechanism, and it takes in a set of scores and is used to choose the object with the highest score. It is typically implemented by adding Laplacian noise to each of the scores and then choosing the one which is largest \cite{mcsherry2007mechanism}. The report-noisy-max mechanism has found wide use in DP literature, and is used by some of the algorithms we review in section III. 

\subsubsection{Global and Local Differential Privacy}

Traditional DP algorithms assume that individuals' raw datapoints are colocated on one machine. Only resulting analyses or models built from the data will be shared publicly with untrusted agents, so only these analyses or models must be noised. This framework is called \textit{global} or \textit{central} DP, in reference to the assumption that all individuals trust one entity to curate their raw data \cite{xiong2020comprehensive}. 

There are a few common methods to achieving DP under the global assumption. The first is output perturbation, in which the output of an algorithm is noised based on its sensitivity. This method is used often with traditional statistical estimators such as the mean, but it can be difficult to calculate the sensitivity of more complicated outputs like machine learning models \cite{rubinstein2009learning, chaudhuri2011differentially}. 

Another method is objective perturbation, which noises the objective function of an optimization problem. This method is flexible to modeling algorithms, but original works required the optimization problem to be solved exactly, which is challenging. More recent frameworks have extended objective perturbation to approximate solutions of optimization problems \cite{chaudhuri2011differentially, kifer2012private}. 

Finally, gradient perturbation privatizes algorithms using gradient descent by noising each step of the optimization process. This method has been shown to work better than output and objective perturbation on many problems, and it is easily modified for the stochastic gradient descent case \cite{bassily2014private}. Its popularity is bolstered by the developments of R\'eyni and zero-concentrated DP, which allow for tight composition of gradient perturbation methods in the non-stochastic and stochastic settings \cite{mironov2017renyi, bun2016concentrated}. 

However, when individuals do not trust a central actor to curate their data, local DP must be used. In this case, individuals noise data prior to sending it to a central server, thus ensuring that curators of the central server cannot re-identify the data's sensitive attributes. A number of noising algorithms exist, but the main drawback of local DP is reduced accuracy and utility \cite{xiong2020comprehensive}. 

To combat reduced accuracy and utility while avoiding trust of the central actor, shuffle differential privacy was developed. In shuffle differential privacy, each individual's data is shuffled prior to being provided to the actor so it cannot be readily linked back to a single user through a specific connection. This allows for reducing the noise added to each individual's data, which can improve accuracy and utility. It can be shown that shuffle differential privacy is a stronger than global differential privacy and weaker than local differential privacy \cite{cheu2021differential}. One common specific application of shuffle differential privacy is aggregate differential privacy, which sums many local differentially private metrics prior to providing this result to an untrusted actor. Aggregate differential privacy has been used in real-world systems for tracking user statistics and application performance \cite{mcmillan2022private}.

In this review, we mention methods for high-dimensional private linear modeling under both the global and local models. However, since methods for high-dimensional linear modeling under local DP are nascent, we only test models built under global DP. Finally, note that we are not aware of any methods which employ shuffle differential privacy for high-dimensional linear models. 

\subsubsection{Sparsity, Stability, and Differential Privacy}

Heuristically, an algorithm which is stable produces similar outputs on similar datasets. Algorithmic stability is a well-studied property in statistics and machine learning as it enables a mathematical analysis of algorithm performance \cite{xu2011sparse}. 

DP algorithms are inherently related to algorithmic stability, as a DP algorithm cannot have a large change in its output when one of its datapoints changes. Indeed, Dwork and Lei formalized this intuition when they studied the relationship between DP and robust statistics \cite{dwork2009differential}. 

Unfortunately, Xu et al. found that sparse algorithms cannot be stable \cite{xu2011sparse}. Applying this to our case of DP, this means that, in some sense, DP algorithms are at odds with sparsity. 

We include this discussion here to highlight the fundamental challenge of producing a sparse DP estimator. While both sparsity and DP are useful tools for data analysis, an effective combination of both is a difficult and open problem. Some of the methods reviewed in this work take steps to resolve this problem. 

\subsection{High-Dimensional Optimization for Linear Models}

In this section, we provide a brief overview of high-dimensional optimizers for linear and logistic regression. Optimization methods for high-dimensional problems are an important open problem, so we focus on well-established methods relevant to section III here. 

First, note that in high-dimensional problems, it is easy for models to overfit the training data. If $n$ is the number of datapoints and $d$ is the number of features, high-dimensional problems have $n < d$, so the number of coefficients in the weight vector exceeds the number of targets \cite{subramanian2013overfitting}. For this reason, we must constrain the weight space to be sparse avoid overfitting \cite{wainwright2019high}. Note that literature on nonprivate optimization often considers sparsity to be a de-facto requirement for high-dimensional problems since it improves utility of solutions greatly; however, as discussed in the prior section, maintaining sparsity with DP can be challenging. 

Two common approaches are used to avoid overfitting in high-dimensional linear models. The first is an $L_0$ contraint on the weight vector, which controls the number of nonzero components in the weight vector. Under this constraint, a linear problem with per-example loss function $\ell(\mathbf{x}_i, y_i; \mathbf{w}_i)$ can be written as 
$$\argmin_{\mathbf{w}:\ \lVert \mathbf{w} \rVert_0 \leq k} \sum_{i = 1}^{N} \ell(\mathbf{x}_i, y_i; \mathbf{w}_i).$$

To solve this problem, iterative gradient hard thresholding can be used. This is a greedy method which retains only the top-$k$ coefficients of the weight vector after each gradient step. Since it is greedy, it is not guaranteed to converge to an optimal solution. However, it approaches the $L_0$ constrained problem in a computationally efficient manner \cite{jain2014iterative}. 

$L_1$ constraints are used more often. In this setting, we solve 
$$\argmin_{\mathbf{w}:\ \lVert \mathbf{w} \rVert_1 \leq k} \sum_{i = 1}^{N} \ell(\mathbf{x}_i, y_i; \mathbf{w}_i).$$
This is a relaxation of the $L_0$ constraint which often produces sparse solutions and can be solved efficiently with a number of optimizers; however, the nonzero components of its solutions are biased towards 0, which can reduce the utility of the output \cite{bertsimas2016best}. 

Solving this problem can be done with a number of optimizers. Most commonly, it is solved with the Frank-Wolfe algorithm, which iteratively takes steps towards the vertex of the $k$-norm $L_1$-ball which has minimum loss \cite{frank1956algorithm}. Compressed learning, in which the dimensionality of the dataset is reduced by multiplication with a random matrix, can also be used \cite{maillard2009compressed}. 

Note that the $L_1$ constrained problem is often written as an $L_1$ regularized problem. Specifically, a linear problem can be written as 
$$\argmin_{\mathbf{w}} \sum_{i = 1}^{N} \ell(\mathbf{x}_i, y_i; \mathbf{w}_i) + \lambda\lVert \mathbf{w} \rVert_1.$$
There is an exact duality between the constrained and regularized setting, where a specific value of $k$ corresponds to a value of $\lambda$. However, the closed form conversion from $k$ to $\lambda$ is not known \cite{boyd2004convex}. 

To solve the regularized problem, coordinate descent can be used. This algorithm iterates through each coefficient of the weight vector, making gradient updates for only that coefficient. More recent algorithms like the alternating direction method of multipliers can also be used, which split the smooth and nonsmooth parts of the regularized optimization problem for better initial optimization \cite{wright2015coordinate,glowinski2014alternating}. 

\begin{figure*}
    \centering    
    \includegraphics[width=0.95\textwidth]{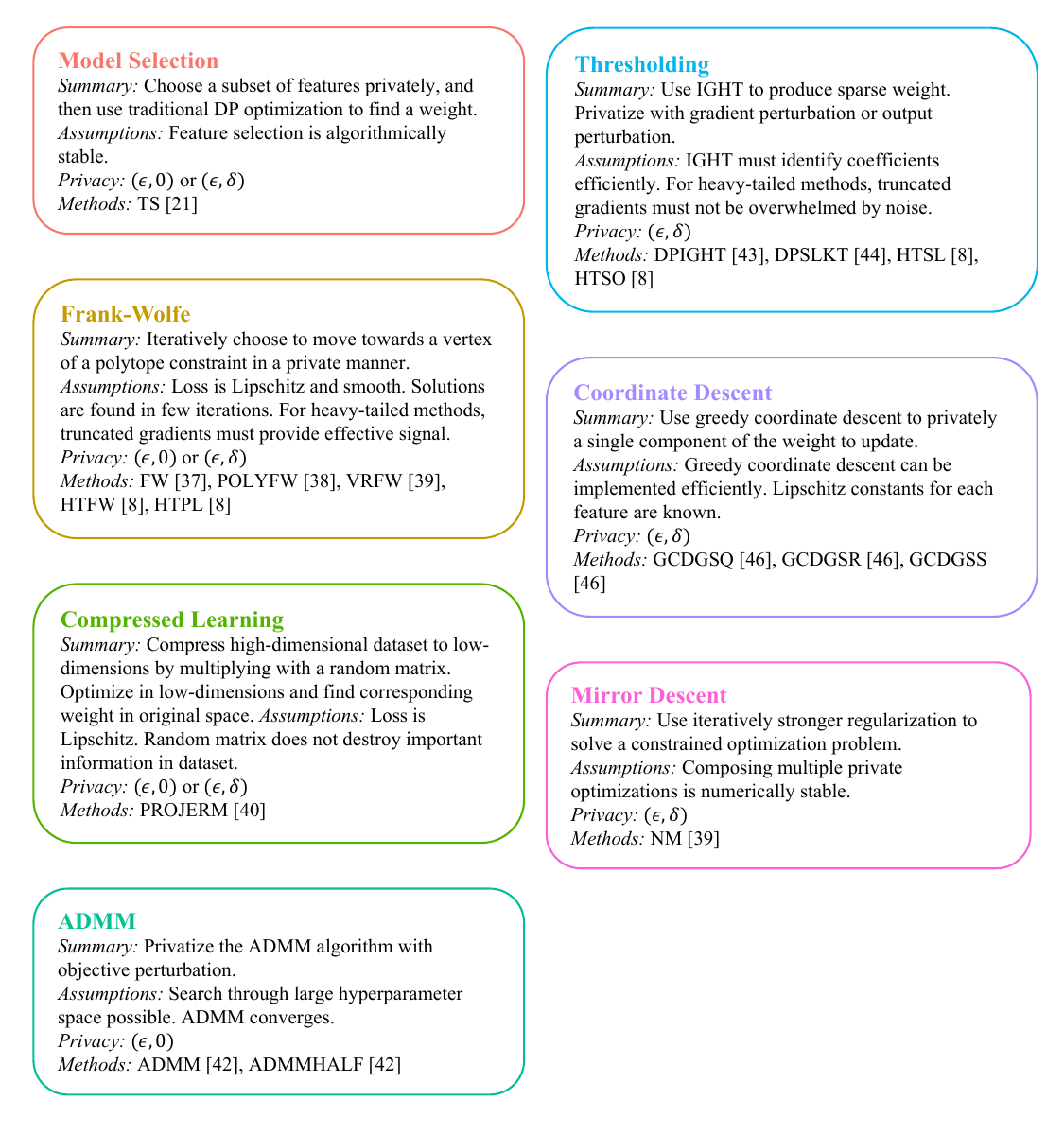}
    \caption{A taxonomy of optimization techniques used for high-dimensional DP linear models.}
\end{figure*}

\section{Review}

This section provides a literature review of algorithms used for high-dimensional DP linear models. The section is organized by optimization technique, and is in roughly chronological order of the first paper considering each technique. Figure 1 provides an overview of these methods, and employs a color-coding we will use in coming sections. Certain optimization techniques have been considered by multiple works, and as such, their subsections are longer than others. Throughout the sections, we highlight potential strengths and weaknesses of each algorithm. These algorithms are then tested empirically in section V. 

In conducting this literature review, we searched for papers considering ``sparse,'' ``high-dimensional,'' ``$L_1$,'' or ``$L_0$'' DP regression. After extensive reading, we identified the works presented in this section as relevant to our review.

Finally, all methods presented in this section employ the unbounded Hamming distance for their adjacency relation. In other words, the definition of differential privacy is employed when $\mathcal{D}$ and $\mathcal{D}'$ differ by one (extra or removed) row. This ensures that the privacy produced by all methods is directly comparable.  

\textcolor[HTML]{F8766D}{\subsection{Model Selection}}

High-dimensional DP linear models were first considered in 2012 by Kifer et al.  \cite{kifer2012private}. Their work is split into two parts: the first developed a generalized objective perturbation mechanism which employs $L_2$ regularization to achieve a strongly-convex loss function. They then derive the utility of their algorithm under both $(\epsilon, 0)$ DP and $(\epsilon, \delta)$ DP, and show that using $(\epsilon, \delta)$ DP saves a factor of $\sqrt{d}$ in utility. 

Their utility analysis shows that this mechanism is not directly applicable to high-dimensional convex optimization. In the second part of their work, they assume that high-dimensional data can be explained by a model with sparsity $s$, and derive algorithms to select a support set of size $s$ prior to running their generalized objective perturbation mechanism. Note that they make a number of additional assumptions when calculating utility, including those on the norm of the dataset and targets along with the restricted strong convexity of the dataset. However, these assumptions are not necessary to guarantee privacy. 

The first algorithm they analyze is derived from the exponential mechanism. Roughly, this method calculates the loss of each support set with size $s$, and uses the exponential mechanism to choose the support set with minimum loss. After choosing the support set, these features are passed into the generalized objective perturbation mechanism. They prove utility of their method when $n = \omega(s^3 \log d)$, but is computationally inefficient since the support selection step must calculate the models of all $d \choose s$ possible support sets. This can be compared to the inefficiency of $L_0$-constrained minimization in the nonprivate setting. 

Their second algorithm uses the sample and aggregate framework to split the training examples into $\sqrt{n}$ blocks and compute the optimal support on each block. After all $\sqrt{n}$ blocks are run, each of the $d$ features has a score associated with it: the number of blocks which chose to include it in their support. A private mechanism can be used to return the top-$s$ features, which are then passed into the generalized objective perturbation mechanism. They prove utility of this method when $n = \omega(s^2 \log^2 d)$, but require an assumption that each of the $\sqrt{n}$ blocks follows the restricted strong convexity assumption, which is strictly stronger than the assumption for the whole dataset. 

Since their method for high-dimensional regression involves two stages, we call it TS. Note that because the method is split into two steps, the privacy budget must also be split between these steps. In their paper, Kifer et al. choose to split the budget into $(\frac{\epsilon}{2}, 0)$ for the support selection step and $(\frac{\epsilon}{2}, \delta)$ for the perturbation mechanism. We follow the same convention  in our experiments. Additionally, our experiments only test the sample and aggregate support selection mechanism since it is computationally efficient. 

Thakurta \& Smith (2013) follow up Kifer et al.'s work by developing DP model selection mechanisms based on perturbation and subsampling stability \cite{thakurta2013differentially}. In short, they call a function $f$ perturbation stable if $f$ outputs the same value on an input dataset $\mathcal{D}$ and all of $\mathcal{D}$'s neighbors. Under this condition, they require an algorithm $\mathcal{A}_{\text{dist}}$ which outputs the distance between $\mathcal{D}$ to its nearest unstable instance. Given the output of $\mathcal{A}_{\text{dist}}$, they use the Propose-Test-Release mechanism to output $f(\mathcal{D})$ in a DP manner. 

However, this mechanism is typically not practical since calculating $\mathcal{A}_{\text{dist}}$ requires searching over all datasets around $\mathcal{D}$, of which there may be infinitely many. Secondly, the mechanism relies on the stability of $f$ at the dataset $\mathcal{D}$, which may change for different functions $f$. 

Thakurta \& Smith then analyze subsampling stable functions. For a dataset $\mathcal{D}$, they define $\widehat{\mathcal{D}}$ as a random subset of $\mathcal{D}$ in which each element appears independently with probability $q$. Then $f$ is called $q$-subsampling stable if $f(\mathcal{D}) = f(\widehat{\mathcal{D}})$ with probability at least $\frac{3}{4}$. Under this condition, they derive an $(\epsilon, \delta)$-DP algorithm which outputs $f(\mathcal{D})$ with high probability when $f$ is $q$-subsampling stable. 

While computationally tractable, this algorithm is still inefficient in that it requires $\mathcal{O}(\frac{1}{q^2})$ runs of $f$. Given the computational complexity of model selection procedures and the fact that $q$ can be very small for low values of $\epsilon$ and $\delta$, this method can be difficult to use in practice. Additionally, there is a nonzero probability that the algorithm outputs $\perp$, a symbol indicating that the model selection procedure cannot be outputted while maintaining DP. Finally, as in the TS approach, using this procedure would require arbitrarily splitting the privacy budget between the model selection and optimization steps. For these reasons, we do not implement this method in our experiments. 

Another potential approach within model selection could involve modifying nonprivate screening rules to operate with DP. Recent work has attempted to do this but has shown a negative result \cite{khanna2023challenge}. It remains an open question whether an effective screening rule can be derived for model selection with high probability. 

\textcolor[HTML]{C49A00}{\subsection{Frank-Wolfe}}

The Frank-Wolfe algorithm (FW) is a greedy optimization technique which works well on problems with polytope constraints. At each iteration, FW finds a first-order approximation of the objective function at the current iterate. It then finds a vertex of the polytope constraint which minimizes the first-order approximation, and takes a step in this direction. The algorithm was first developed for quadratic programming, but can be efficiently implemented for many loss functions including mean-squared error and binary cross-entropy. 

Talwar et al. (2015) privatize FW for $L_1$ constrained optimization \cite{talwar2015nearly}. The algorithm works well in this setting since the $L_1$ constraint corresponds to a polytope. In order to produce private FW, at each iteration they treat the values of the first-order approximation at each vertex as scores, and use the exponential mechanism to choose a direction to move in. Using the advanced composition theorem for $(\epsilon, 0)$ DP, they can compute the exact amount of noise to add, achieving $(\epsilon, \delta)$ DP. 

Their utility analysis does not require restricted strong convexity or restricted strong smoothness. Instead, it relies on the curvature constant of the loss function, which is bounded for both linear and logistic regression. This constant and with the Lipschitzness of the loss function with respect to the $L_1$ norm is the only information necessary to produce the utility bound. 

Due to the lack of assumptions in producing a utility bound, private FW was explored by a number of subsequent works. These works identified that private FW requires gradients with respect to the entire dataset at each iteration, which can waste privacy budget. According to the principles of subsampling in DP, if the entire dataset is not accessed at each iteration, less information from the data is used in each update, and less noise is required for privacy. The two following methods consider stochastic private FW methods. 

Bassily et al. (2021) consider a stochastic private FW method which reduces variance of updates through a recursive gradient estimator \cite{bassily2021non}. Specifically, their method computes the gradient of the objective function with respect to $\frac{n}{2}$ of the datapoints. Then it iterates through the remaining $\frac{n}{2}$ points and updates the gradient with a weighted average. The weight is updated after each iteration in a private manner using the exponential mechanism, and advanced composition yields $(\epsilon, \delta)$-DP. Using the author's convention, we call this method POLYFW. 

POLYFW has error of $\widetilde{\mathcal{O}}(\frac{\log d}{\epsilon \sqrt{n}})$, which is not optimal according to the work below. Their analysis does not explicitly rely on the curvature constant but rather requires $L_1$ Lipschitzness and a smooth loss function with respect to the $L_1$ norm. 

Asi et al. (2021) develop a variance-reduced FW method (VRFW) for smooth functions using DP binary trees \cite{asi2021private}. Nodes in these trees closer to the root are assigned more samples, making their gradient more accurate, while nodes closer to the leaves have less samples. Iterating over the leaves, their algorithm makes FW updates which consider the gradients collected over the entire path from the root to the leaf. By reusing gradients calculated from more datapoints, the algorithm is able to add less noise through the exponential mechanism. Their method can achieve both $(\epsilon, 0)$-DP and $(\epsilon, \delta)$-DP. 

Asi et al. calculate the error of this method, which is $\widetilde{\mathcal{O}}(1/\sqrt{n} + 1/(n\epsilon)^{2/3})$ for $(\epsilon, \delta)$-DP. They then calculate lower bounds of error, and show that this error matches the lower bound. For this reason, their algorithm has optimal utility for smooth loss functions. 

Finally, Hu et al. (2022) develop two stochastic FW-based algorithms for heavy tailed high-dimensional data \cite{hu2022high}. Each of their utility analyses assumes bounded fourth order moments, which is common in heavy tailed statistics. 

The first method they use involves a soft truncation and scaling of the gradient. They do this because heavy-tailed data may not have bounded gradients, which can lead to unbounded sensitivity. With these truncated gradients, the exponential mechanism is used to choose a direction for minimization. This algorithm is $(\epsilon, 0)$-DP. We call this method HTFW. 

When specifically considering linear regression, they find that truncating each element of the design matrix and target vector to be in $[-K, K]$ can bound sensitivity. After this data processing step, they use the exponential mechanism to choose a direction for minimization. This algorithm is $(\epsilon, \delta)$-DP, and Hu et al. abbreviate it as HTPL. 

\textcolor[HTML]{53B400}{\subsection{Compressed Learning}}

Another approach to constrained optimization is through compressed learning, which reduces the dimensionality of the input space by multiplying the design matrix by a random matrix $\bm{\Phi} \in \mathbb{R}^{m \times d}$. Typically $\bm{\Phi}$ is chosen to be a subgaussian random matrix with norm $\mathcal{O}(1)$ and the feasible set $\mathcal{C}$ is the scaled $L_1$ ball. 

Kasiviswanathan \& Jin (2016) employ compressed sensing for high-dimensional DP linear modelling \cite{kasiviswanathan2016efficient}. After choosing $\bm{\Phi}$ and $\mathcal{C}$, they solve the problem $$\argmin_{\vartheta \in \bm{\Phi}\mathcal{C}} \frac{1}{n} \sum_{i = 1}^{n} \ell (\langle \Phi\mathbf{x}_i, \vartheta \rangle; y_i)$$ with a DP optimizer. To obtain an estimate for the parameter vector in the data space, they solve $$\theta \in \argmin_{\theta \in \mathbb{R}^d: \bm{\Phi}\theta = \vartheta } \lVert \theta \rVert_{\mathcal{C}}$$ where $\lVert \cdot \rVert_\mathcal{C}$ represents the Minkowski norm induced by the feasible set. When $\mathcal{C}$ is the scaled $L_1$ ball, this is equivalent to the $L_1$ norm. Note that the output of this procedure is DP with the same parameters as $\vartheta$ since $\theta$ is only a function of $\bm{\Phi}$, which does not depend on the data, and $\vartheta$, which is DP. 

Since their procedure only requires a DP optimizer to guarantee DP, it can be both $(\epsilon, 0)$-DP and $(\epsilon, \delta)$-DP. In our experiments, we implement an $(\epsilon, \delta)$-DP optimizer to work with their procedure. We refer to their procedure as PROJERM, in reference to its projected empirical risk minimization step. 

The utility bounds which Kasiviswanathan \& Jin derive follow directly from those known from the Johnson-Lindenstrauss lemma. Their bounds hold under conditions on size of $m$. 

Zheng et al. (2017) used compressed learning to develop an algorithm for high-dimensional linear regression under local DP \cite{zheng2017collect}. Instead of using gradient perturbation like Kasiviswanathan \& Jin, Zheng et al. noise the projection of each datapoint $\bm{\Phi} \mathbf{x}_i$ prior to optimization. Using the noised datapoints, they correct the quadratic loss function of linear regression to ensure it is unbiased, and optimize over this corrected loss function. Their algorithm is $(\epsilon, \delta)$-DP. 

Zheng et al. use matrix concentration inequalities to derive bounds on utility in the value of the loss. However, they make no claims in directly comparing the output $\mathbf{w}_{\text{priv}}$ from their algorithm to the optimal $\mathbf{w}^*$. 

\textcolor[HTML]{00C094}{\subsection{ADMM}}

The alternating direction method of multipliers (ADMM) is a common algorithm used for optimization of nonsmooth or nonconvex regularizers by transforming an optimization problem into two simpler problems. Under assumptions on the strong convexity of the loss function and separability of the regularizer, ADMM can be shown to converge to a stationary point. 

Wang \& Zhang (2020) use the ADMM algorithm and use it to achieve DP $L_1$ and $L_{1/2}$ regularized logistic regression (ADMM and ADMMHALF) \cite{wang2020differential}. First, they show that these methods access the data in only the second of its three steps. Then, to privatize the algorithm, they employ objective perturbation in every second step. Finally, they use the basic composition theorem to show that with appropriate noise, their algorithm is $(\epsilon, 0)$-DP after $K$ iterations. 

Although Wang \& Zhang do not provide a utility analysis of ADMM and ADMMHALF, they do analyze the three steps of each iteration and find that only the first step is unstable with sparsity. Since the data is only accessed in the second step, privatizing the second step does not require excessive noise since this step is convex and stable. For this reason, they intuitively claim that their algorithm is well-suited for DP, which outputs accurate results on stable functions. 

\textcolor[HTML]{00B6EB}{\subsection{Thresholding}}

A simple approach to sparse optimization is to enforce sparsity throughout training. This can be done with hard thresholding operations, in which only the top $s$ components of a partially learned weight are retained after each optimizer step. Hard thresholding based approaches have been considered in many forms for high-dimensional DP optimization. We cover these methods below. 

Wang \& Gu (2019) consider DP iterative gradient hard thresholding (DPIGHT), which retains the top $s$ components of parameter vector after each gradient step \cite{wang2019differentially}. Although their method is simple, when they assume that each row of the design matrix is sub-Gaussian, they achieve the same utility as Kifer et al. in the TS procedure without requiring an extra support selection step. However, their utility does assume the sparse eigenvalue condition, which implies both restricted strong convexity and restricted strong smoothness. They analyze the privacy of their algorithm with respect to zero-concentrated DP as this allows for tight composition of gradient descent. Note that zero-concentrated DP can easily be converted to $(\epsilon, \delta)$-DP. 

Wang \& Gu (2020) follow up this work by considering a DP knowledge transfer framework (DPSLKT) to produce sparse linear models \cite{wang2020knowledge}. First, they use IGHT to train a teacher model on private data without DP. Next, they generate an auxiliary training set and use output perturbation when passing it through the initially trained model. Now this dataset is DP. Finally, to produce a sparse model, they use IGHT to find a student model which fits the the DP data well. Due to the post-processing property of DP, this model is also DP. Since this method also uses gradient descent, Wang \& Gu use zero-concentrated DP, which can be converted to $(\epsilon, \delta)$-DP. 

The utility guarantees of DPSLKT outperform those of previous methods because they rely on the $L_\infty$ norm of the data, which can be up to $\mathcal{O}(\sqrt{d})$ times less than the $L_2$ norm. However, their utility analysis requires restricted strong convexity and restricted strong smoothness. For this reason, Wang \& Gu give an option to define the loss function with an $L_2$ regularization term, since this will ensure that the loss is strongly convex. This approach is similar to the TS method of Kifer et al. However, unlike the TS method, this regularization term is not necessary, but without it, an assumption of restricted strong convexity and restricted strong smoothness is necessary, which typically does not hold on real-world datasets. 

Hu et al. (2022) employ iterative hard thresholding to develop a private linear regression algorithm for heavy tailed high-dimensional data, which they call HTSL \cite{hu2022high}. To do this, they truncate each element of the design matrix and target vector to be in $[-K, K]$, where $K$ is a value set by the user. Then they use gradient descent for optimization, but after each step they privately retain the top $s$ elements of the weight vector. Since their data and targets are truncated, they can calculate the sensitivity of this step, which is passed into the exponential mechanism. Using the advanced composition theorem, they achieve $(\epsilon, \delta)$-DP. 

Note that truncation is required because the data is heavy tailed. Heavy tailed data is typically not bounded or subgaussian, and its loss function is typically not $\mathcal{O}(1)$ Lipschitz. Some combintation of these assumptions are required to compute the sensitivity for non-heavy tailed algorithms. To avoid this, Hu et al. truncate the ``outlier'' elements in the design matrix and target vector. However, the utility bound which Hu et al. derive requires that the dataset and targets have bounded fourth order moment. This assumption is similar to those made by other works in heavy tailed optimization. 

For general convex optimization, Hu et al. use a soft trunctation and scaling operation on the gradient. Then they use gradient descent, privately retaining the top $s$ elements of the weight vector. In this case, they bound the optimization sensitivity through the truncation and scaling of the gradient. Again, their utility bound requires bounded fourth order moments. This method produces and $(\epsilon, \delta)$-DP result. They call this method HTSO, since it performs heavy-tailed sparse optimization. 

Finally, we mention Wang \& Xu's 2019 work on iterative hard thresholding methods for local DP \cite{wang2019sparse}. In this work, they assume that each datapoint $\mathbf{x}_i$ is sampled from $\{-1, +1\}^d$, which is a very strong condition. 

However, this work is valuable since it shows that even under such strong conditions, the utility must have order $\widetilde{\Omega}(d\log d)$. Their iterative hard thresholding algorithm achieves this, but in general, orders greater than $\widetilde{\Omega}(\log d)$ are considered unlearnable in the high-dimensional setting. They then demonstrate that if only the labels $y_i$ are kept private, iterative hard thresholding can achieve a bound of $\widetilde{\mathcal{O}}(\log d)$. However, this setting is dissimilar from the DP setting we consider in this paper, and thus we do not implement this algorithm. 

\textcolor[HTML]{A58AFF}{\subsection{Coordinate Descent}}

Coordinate descent is an optimization algorithm which iteratively updates only one parameter of a model's weight using its gradient. The algorithm typically chooses parameters to update cyclically or at random, but greedy coordinate descent chooses to update the parameter with the highest absolute gradient. In the nonprivate setting, all of these methods perform well on high-dimensional linear problems. Non-greedy methods are typically used since they are computationally efficient, only requiring the gradient with respect to one parameter. 

Mangold et al. (2022) develop a DP coordinate descent algorithm for high-dimensional ERM \cite{mangold2022high}. They choose to use greedy coordinate descent (GCD), which updates the coordinate with the largest absolute value gradient, because under DP fewer computations enable less added noise. They acknowledge that this is an inherent tradeoff to computational efficiency, since their procedure must compute the gradient with respect to every parameter at every iteration. 

Since they update only one parameter per iteration, Mangold et al. are able to employ per-feature smoothness and Lipschitz constants. In some datasets, some features can have significantly lower smoothness and Lipschitz constants as compared to the entire dataset. For these features, less noise is added during coefficient selection and optimization. Finally, using the advanced composition theorem, their method achieves $(\epsilon, \delta)$-DP. 

Their utility bound demonstrates that their algorithm converges to a ball around the optimal weight, where the radius of the ball is determined by the level of privacy. However, it is worth noting that their utility bounds consider only smooth loss functions, and are thus not applicable to $L_1$ regularizers. To use $L_1$ regularizers, Mangold et al. develop DP proximal coordinate descent algorithms with a variety of proximal operators. These algorithms are named GCDGSQ, GCDGSR, and GCDGSS, to represent three different proximal operators, and empirical results indicate that these modifications greatly improve performance on high-dimensional datasets. However, they do not consider the utility of the algorithm with proximal operators. 

\textcolor[HTML]{FB61D7}{\subsection{Mirror Descent}}

Asi et al. (2021) privatize an iterative localization framework with private optimization. They call their method private mirror descent with noise multipliers (NM) \cite{asi2021private}. This method finds the optimal parameter vector for a sequence of optimization problems with increasing regularization. By sequentially increasing the regularization, the method finds iteratively smaller regions for optimization, making the optimization problem easier. In addition, the regularization makes the optimization problem strongly convex, making it easy to use well-known optimal optimization techniques for private strongly convex problems. They prove that NM achieves optimal utility for non-smooth loss functions by demonstrating that using private optimization with iterative localization produces a loss which matches a known lower bound up to logarithmic factors. This algorithm is $(\epsilon, \delta)$-DP.

\vspace{1em}

\section{Implementation Details}

In the following section, we perform experiments on the central DP methods previously discussed. We release all code for these experiments online. This is one of the major contributions of this work, as previous works do not release code or test algorithms empirically. While theoretical utility bounds can provide an even ground for algorithm comparison, these bounds often make significant (and different) assumptions; empirical performance on a variety of datasets better demonstrates how these algorithms would perform if used in application. In this section, we list challenges we faced when implementing some of these methods. 

\begin{itemize}
    \item Without an optimized linear algebra implementation, the FW algorithm is very inefficient. An optimized implementation can be found in \cite{khanna2023sparse}. This holds for the other variants of FW as well. 
    \item VRFW required smoothness constants to be greater than a threshold. Since a $k$-smooth function is also $l$-smooth when $l \geq k$, when implementing this method we set the smoothness constant to $\text{max}(k, l)$ where $k$ is the smoothness constant of our loss function and $l$ is the required lower bound on smoothness.
    \item For heavy-tailed methods requiring the robust gradient, the expression of the correction factor can be found in Lemma 3.2 in \cite{catoni2017dimension}. The authors considering heavy-tailed private optimizers claimed to provide this expression in their appendix, but we could not find it. 
    \item The private optimization method recommended by the authors of PROJERM requires $n^2$ steps of stochastic gradient descent, where $n$ is the number of datapoints. Additionally, the method requires solving a optimization problem in $d$ variables. We did not modify these steps, but for larger datasets PROJERM did not converge within 48 hours. 
    \item ADMM and ADMMHALF have a significant number hyperparamters and can be unstable for many hyperparameters and datasets. We had to refactor portions of their algorithm into numerically stable mathematical equivalents. Refactoring improved performance but certain datasets were still unstable. 
    \item We did not implement GCD methods but instead used code released with the GCD paper to run these methods. 
    \item When implementing NM, we tried to use CVXPY to solve each optimization problem \cite{diamond2016cvxpy,agrawal2018rewriting}. However, the constraints in this problem are unstable, and the free ECOS solver included in CVXPY would produce errors prior to convergence \cite{domahidi2013ecos}. The free SCS solver would take up to 10 seconds for a single iteration, and the cumulative sequence of problems would require over 10,000 iterations, making it infeasible \cite{ocpb:16}. For this reason, we approximated the solution to each optimization problem using gradient descent.
    \item We employed differentially private stochastic gradient descent (DPSGD) as a baseline for each dataset \cite{abadi2016deep}. In addition to being a common DP optimization method, works have shown that the noise added during DPSGD can be an effective regularization strategy which prevents overfitting. This may cause it to perform well in high-dimensional settings \cite{tobaben2023efficacy}. 
\end{itemize}

Finally, it is important to note that all experiments were conducted with Intel(R) Xeon(R) CPU E5-2698 v4 @ 2.20GHz. For each dataset-algorithm combination, 20 trials were performed on different CPUs to speed up training. A 48 hour wall-clock time limit on computation time was set for each dataset-algorithm combination.

\section{Experiments}

This section runs experiments on six datasets to test the methods described in Section III. These experiments provide a systematic comparison between algorithms' performance on varying privacy budgets. Our study focuses on two of the most prevalent linear models: linear regression and logistic regression. 

All datasets were chosen from the libsvm and OpenML libraries \cite{CC01a,OpenML2013}. For linear regression, we used the Bodyfat, PAH, and E2006 datasets. The raw datasets for Bodyfat and PAH were used, but for computational efficiency, 500 datapoints from the E2006 dataset were chosen and their dimensionality was reduced to 500 with PCA. Similarly, the Heart, DBworld-subjects-stemmed, and RCV1 datasets were used for logistic regression. We used the raw Heart and DBworld datasets, but chose 500 datapoints of the RCV1 dataset and reduced their dimensionality to 500 with PCA. These datasets were chosen to identify algorithms' performance on different scales of dimensionality while retaining computational efficiency. Table I provides a summary of the datasets' dimensionalities. 

All features in datasets were demeaned, and then samples in each dataset were rescaled so the maximum $L_1$-norm of a sample in any given dataset was $1$. This was done to bound the Lipschitz and smoothness constants of the datasets under linear and logistic regression, a common requirement in DP optimization literature and detailed well in \cite{kasiviswanathan2016efficient}. Finally, linear or logistic regression models were trained on the datasets under a variety of hyperparameters. Each set of hyperparameters was tested for 20 trials, with a unique 60\%/20\%/20\% split between training, validation, and testing datasets used for each trial. A detailed list of the hyperparameters used for each algorithm can be found in the appendix.

Note that algorithms were tested with privacy parameters $\epsilon \in \{0.1, 0.5, 1, 2, 5\}$ with $\delta = \frac{1}{n_\text{train}^2}$, where $n_\text{train}$ is the number of training samples. These privacy parameters were chosen according to common recommendations in DP literature. Finally, each algorithm was run with $\epsilon = 100$ and $\delta = 0.999$ to approximate a nonprivate solution. We did this for each algorithm so that we find the best nonprivate solution using the same choices of hyperparameters used for private training.

\begin{table}[]
    \caption{Summary of Datasets}
    \label{tab:my_label}
    \centering
    \begin{small}
    \begin{sc}
    \SetTblrInner{rowsep=0pt}
    \begin{tblr}{lccc}
    \toprule
    Dataset & N & D & Regression Type \\ 
    \midrule
    Bodyfat & 252 & 14 & Linear \\ 
    PAH & 80 & 112 & Linear \\ 
    E2006 & 500 & 500 & Linear \\ 
    Heart & 270 & 13 & Logistic \\ 
    DBworld & 64 & 229 & Logistic \\ 
    RCV1 & 500 & 500 & Logistic \\
    \bottomrule
    \end{tblr}
    \end{sc}
    \end{small}
\end{table}

\subsection{Linear Regression}

Table II, Table III, and Table IV demonstrate the mean of 20 trials of mean-squared errors for the test portions of the Bodyfat, PAH, and E2006 datasets, respectively. Additionally, subscripts designating two times the standard error of the mean are provided. For each value of $\epsilon$, the best performing algorithm's results are bolded. Note that ProjERM is not included in Table IV since it did not converge within 48 hours. We now discuss notable performance trends which are seen in these tables. Note that in the appendix, Figure 2, Figure 3, and Figure 4 provide boxplots of the mean-squared errors for the trials on each dataset. 

Perhaps the most significant trend among all three tables is that HTSO, or heavy-tailed sparse optimization, performs well on all values of $\epsilon$. HTSO performs iterative gradient hard thresholding using the robust gradient of the data. Intuitively, this makes sense; while most other methods tested require the loss to be Lipschitz, this constant is often a worst-case bound. For example, although bounding the Lipschitz constant of the loss requires rescaling the datapoints, datapoints with larger rescaled norms might be outliers, and most datapoints have norms near $0$. In contrast, heavy tailed optimizers do not assume Lipschtizness of the loss, and instead use parameters to calculate a robust gradient which is resistant to outliers. Thus, even in settings where the Lipschitz constant of the loss can be set, heavy-tailed private gradient optimizers can perform better. 

Next, we note that GCD, or greedy coordinate descent, can perform also well. Similar to HTSO, these methods do not rely on the Lipschitz constant for optimization of the entire weight vector but rather use different Lipschitz constants for each coefficient. For this reason, private optimization can add less noise to coefficients which are smaller than others, thus improving the utility of the method. Finally, as was reported in \cite{mangold2022high}, different greedy selection rules can perform very differently depending on the dataset. 

Note that GCD methods can be made to better approximate the behavior of HTSO through a clipping parameter. Specifically, by aggressively clipping gradients before applying coordinate descent, GCD can resist outlier influence on gradients. However, our goal was to observe the performance of algorithms without additional modifications, and we did not try to use significant clipping here. 

\noindent\fbox{\parbox{\linewidth}{
\hspace{0.30em} From these results, we identify that future research on private statistical models which incorporate measures of scale and robustness could be particularly promising. Specifically, robust calculations are statistically stable, which reduces the noise required for DP. Considering the scale of features in a linear model allows the information from features ``squashed'' after normalization to not be lost. 
}}

\begin{table*}[]
    \caption{\textbf{Bodyfat}: Mean Squared Error}
    \label{tab:my_label}
    \centering
    \begin{small}
    \begin{sc}
    \SetTblrInner{rowsep=0pt}
    \begin{tblr}{lcccccc}
    \toprule
    $\epsilon$ &                0.1   &                0.5   &                1.0   &                2.0   &                5.0   &                Nonprivate \\
    \midrule
    \textcolor[HTML]{F8766D}{ts}      &  $0.1142_{(0.0067)}$ &  $0.1133_{(0.0066)}$ &  $0.1122_{(0.0065)}$ &  $0.1103_{(0.0064)}$ &   $0.1051_{(0.0060)}$ &   $\mathbf{0.0112}_{(0.0070)}$ \\
    \hline[dashed]
    \textcolor[HTML]{C49A00}{fw}      &  $0.0947_{(0.0053)}$ &  $0.0947_{(0.0054)}$ &  $0.0947_{(0.0054)}$ &  $0.0947_{(0.0054)}$ &  $0.0947_{(0.0054)}$ &  $0.0844_{(0.0047)}$ \\
    \textcolor[HTML]{C49A00}{polyfw}  &  $0.0955_{(0.0053)}$ &  $0.0955_{(0.0053)}$ &  $0.0955_{(0.0053)}$ &  $0.0955_{(0.0053)}$ &  $0.0955_{(0.0053)}$ &  $0.0955_{(0.0053)}$ \\
    \textcolor[HTML]{C49A00}{vrfw}    &  $0.0965_{(0.0057)}$ &  $0.0965_{(0.0057)}$ &  $0.0965_{(0.0057)}$ &  $0.0965_{(0.0057)}$ &  $0.0965_{(0.0057)}$ &  $0.0892_{(0.0058)}$ \\
    \textcolor[HTML]{C49A00}{htfw}    &  $0.0946_{(0.0053)}$ &  $0.0946_{(0.0053)}$ &  $0.0946_{(0.0053)}$ &  $0.0937_{(0.0058)}$ &  $0.0896_{(0.0049)}$ &  $0.0886_{(0.0049)}$ \\
    \textcolor[HTML]{C49A00}{htpl}    &  $0.0946_{(0.0053)}$ &  $0.0946_{(0.0053)}$ &  $0.0946_{(0.0053)}$ &  $0.0946_{(0.0053)}$ &  $0.0946_{(0.0053)}$ &  $0.0847_{(0.0047)}$ \\
    \hline[dashed]
    \textcolor[HTML]{53B400}{projerm} &  $0.0945_{(0.0053)}$ &  $0.0945_{(0.0053)}$ &  $0.0945_{(0.0053)}$ &  $0.0945_{(0.0053)}$ &  $0.0945_{(0.0053)}$ &   $0.0890_{(0.0051)}$ \\
    \hline[dashed]
    \textcolor[HTML]{00B6EB}{dpight}  &  $0.0891_{(0.0052)}$ &  $0.0926_{(0.0054)}$ &  $0.0896_{(0.0055)}$ &  $0.0892_{(0.0056)}$ &  $0.0916_{(0.0052)}$ &  $0.0932_{(0.0053)}$ \\
    \textcolor[HTML]{00B6EB}{dpslkt}  &   $0.0902_{(0.0060)}$ &   $0.0902_{(0.0060)}$ &   $0.0902_{(0.0060)}$ &   $0.0902_{(0.0060)}$ &  $0.0908_{(0.0052)}$ &  $0.0744_{(0.0075)}$ \\
    \textcolor[HTML]{00B6EB}{htsl}    &  $0.0896_{(0.0065)}$ &  $0.0885_{(0.0059)}$ &  $0.0894_{(0.0059)}$ &  $0.0886_{(0.0054)}$ &  $0.0886_{(0.0056)}$ &  $0.0942_{(0.0055)}$ \\
    \textcolor[HTML]{00B6EB}{htso}    &  $\mathbf{0.0685}_{(0.0091)}$ &  $0.0727_{(0.0153)}$ &  $0.0699_{(0.0458)}$ &  $\mathbf{0.0521}_{(0.0131)}$ &  $0.0607_{(0.0104)}$ &  $0.0729_{(0.0068)}$ \\
    \hline[dashed]
    \textcolor[HTML]{A58AFF}{gcdgsq}  &  $1.0307_{(0.1095)}$ &  $0.0768_{(0.0082)}$ &  $0.0661_{(0.0046)}$ &  $0.0701_{(0.0046)}$ &  $0.0755_{(0.0057)}$ &  $0.0162_{(0.0022)}$ \\
    \textcolor[HTML]{A58AFF}{gcdgsr}  &  $0.0956_{(0.0053)}$ &  $\mathbf{0.0041}_{(0.0018)}$ &  $\mathbf{0.0312}_{(0.0024)}$ &  $0.0582_{(0.0033)}$ &  $0.0697_{(0.0053)}$ &  $0.0615_{(0.0064)}$ \\
    \textcolor[HTML]{A58AFF}{gcdgss}  &  $0.0956_{(0.0053)}$ &  $0.0779_{(0.0048)}$ &  $0.0894_{(0.0051)}$ &  $0.0956_{(0.0069)}$ &   $\mathbf{0.0462}_{(0.0090)}$ &  $0.0435_{(0.0042)}$ \\
    \hline[dashed]
    \textcolor[HTML]{FB61D7}{nm}      &  $0.0956_{(0.0053)}$ &  $0.0956_{(0.0053)}$ &  $0.0953_{(0.0053)}$ &  $0.0973_{(0.0053)}$ &  $0.0948_{(0.0054)}$ &   $0.0980_{(0.0052)}$ \\
    \hline[dashed]
    sgd      &  $0.0916_{(0.0051)}$ &  $0.0916_{(0.0051)}$ &  $0.0916_{(0.0051)}$ &  $0.0916_{(0.0051)}$ &  $0.0916_{(0.0051)}$ &  $0.0983_{(0.0055)}$ \\
    \bottomrule
    \end{tblr}
    \end{sc}
    \end{small}
\end{table*}

\begin{table*}[]
    \caption{\textbf{PAH}: Mean Squared Error}
    \label{tab:my_label}
    \centering
    \begin{small}
    \begin{sc}
    \SetTblrInner{rowsep=0pt}
    \begin{tblr}{lcccccc}
    \toprule
    $\epsilon$ &                      0.1   &                   0.5   &                   1.0   &                2.0   &                5.0   &                Nonprivate \\
    \midrule
    \textcolor[HTML]{F8766D}{ts}      &        $0.0607_{(0.0252)}$ &   $0.0572_{(0.0245)}$ &    $0.0607_{(0.0257)}$ &  $0.0608_{(0.0262)}$ &  $0.0611_{(0.0275)}$ &  $0.0837_{(0.0224)}$ \\
    \hline[dashed]
    \textcolor[HTML]{C49A00}{fw}      &         $0.0910_{(0.0168)}$ &    $0.0910_{(0.0168)}$ &     $0.0910_{(0.0168)}$ &   $0.0910_{(0.0168)}$ &   $0.0910_{(0.0168)}$ &  $0.0902_{(0.0166)}$ \\
    \textcolor[HTML]{C49A00}{polyfw}  &        $0.0909_{(0.0168)}$ &   $0.0909_{(0.0168)}$ &    $0.0909_{(0.0168)}$ &  $0.0909_{(0.0168)}$ &  $0.0909_{(0.0168)}$ &  $0.0911_{(0.0168)}$ \\
    \textcolor[HTML]{C49A00}{vrfw}    &         $0.0918_{(0.0170)}$ &    $0.0918_{(0.0170)}$ &     $0.0918_{(0.0170)}$ &   $0.0918_{(0.0170)}$ &   $0.0918_{(0.0170)}$ &  $0.0911_{(0.0169)}$ \\
    \textcolor[HTML]{C49A00}{htfw}    &        $0.0907_{(0.0167)}$ &   $0.0907_{(0.0167)}$ &    $0.0907_{(0.0167)}$ &  $0.0907_{(0.0167)}$ &  $0.0905_{(0.0167)}$ &   $0.0910_{(0.0168)}$ \\
    \textcolor[HTML]{C49A00}{htpl}    &        $0.0907_{(0.0167)}$ &   $0.0908_{(0.0168)}$ &    $0.0907_{(0.0167)}$ &  $0.0907_{(0.0167)}$ &  $0.0907_{(0.0167)}$ &  $0.0905_{(0.0167)}$ \\
    \hline[dashed]
    \textcolor[HTML]{53B400}{projerm} &         $0.0910_{(0.0168)}$ &    $0.0910_{(0.0168)}$ &     $0.0910_{(0.0168)}$ &   $0.0910_{(0.0168)}$ &   $0.0910_{(0.0168)}$ &  $0.0911_{(0.0168)}$ \\
    \hline[dashed]
    \textcolor[HTML]{00B6EB}{dpight}  &        $0.0892_{(0.0163)}$ &   $0.0897_{(0.0164)}$ &    $0.0892_{(0.0163)}$ &  $0.0904_{(0.0167)}$ &  $0.0904_{(0.0166)}$ &   $0.0910_{(0.0168)}$ \\
    \textcolor[HTML]{00B6EB}{dpslkt}  &        $\mathbf{0.0538}_{(0.0068)}$ &    $0.0850_{(0.0146)}$ &    $0.0538_{(0.0068)}$ &  $0.0852_{(0.0138)}$ &   $0.0850_{(0.0146)}$ &   $0.0828_{(0.0170)}$ \\
    \textcolor[HTML]{00B6EB}{htsl}    &        $0.0905_{(0.0168)}$ &   $0.0901_{(0.0166)}$ &    $0.0903_{(0.0166)}$ &  $0.0905_{(0.0167)}$ &  $0.0908_{(0.0168)}$ &  $0.0911_{(0.0168)}$ \\
    \textcolor[HTML]{00B6EB}{htso}    &        $0.0581_{(0.0157)}$ &   $\mathbf{0.0352}_{(0.0063)}$ &    $\mathbf{0.0432}_{(0.0062)}$ &   $\mathbf{0.0530}_{(0.0077)}$ &  $0.0647_{(0.0102)}$ &  $0.0909_{(0.0168)}$ \\
    \hline[dashed]
    \textcolor[HTML]{A58AFF}{gcdgsq}  &    $429.1658_{(195.0152)}$ &  $15.6425_{(4.2077)}$ &    $3.4062_{(0.9211)}$ &  $1.1643_{(1.2433)}$ &  $\mathbf{0.0591}_{(0.0231)}$ &  $0.0474_{(0.0076)}$ \\
    \textcolor[HTML]{A58AFF}{gcdgsr}  &  $1366.9646_{(1497.3976)}$ &  $56.6219_{(61.293)}$ &  $38.2438_{(49.2854)}$ &   $1.5510_{(0.8699)}$ &  $0.1607_{(0.0751)}$ &  $0.0855_{(0.0169)}$ \\
    \textcolor[HTML]{A58AFF}{gcdgss}  &        $1.4834_{(0.9205)}$ &   $0.2219_{(0.0879)}$ &    $0.1485_{(0.0459)}$ &  $0.1188_{(0.0301)}$ &  $0.1179_{(0.0522)}$ &  $\mathbf{0.0395}_{(0.0065)}$ \\
    \hline[dashed]
    \textcolor[HTML]{FB61D7}{nm}      &        $0.0911_{(0.0168)}$ &   $0.0911_{(0.0168)}$ &    $0.0911_{(0.0168)}$ &  $0.0911_{(0.0168)}$ &  $0.0911_{(0.0168)}$ &  $0.0911_{(0.0168)}$ \\
    \hline[dashed]
    sgd      &        $0.0914_{(0.0170)}$ &   $0.0914_{(0.0170)}$ &    $0.0914_{(0.0170)}$ &  $0.0914_{(0.0170)}$ &  $0.0914_{(0.0170)}$ &  $0.0910_{(0.0168)}$ \\
    \bottomrule
    \end{tblr}
    \end{sc}
    \end{small}
\end{table*}

\begin{table*}[]
    \caption{\textbf{E2006}: Mean Squared Error}
    \label{tab:my_label}
    \centering
    \begin{small}
    \begin{sc}
    \SetTblrInner{rowsep=0pt}
    \begin{tblr}{lcccccc}
    \toprule
    $\epsilon$ &                      0.1   &                   0.5   &                   1.0   &                2.0   &                5.0   &                Nonprivate \\
    \midrule
    \textcolor[HTML]{F8766D}{ts}      &  $0.0392_{(0.0033)}$ &  $0.0393_{(0.0033)}$ &  $0.0393_{(0.0033)}$ &  $0.0393_{(0.0033)}$ &   $0.0380_{(0.0035)}$ &  $0.0444_{(0.0052)}$ \\
    \hline[dashed]
    \textcolor[HTML]{C49A00}{fw}      &   $0.0390_{(0.0033)}$ &   $0.0390_{(0.0033)}$ &   $0.0390_{(0.0033)}$ &   $0.0390_{(0.0033)}$ &   $0.0390_{(0.0033)}$ &  $\mathbf{0.0382}_{(0.0033)}$ \\
    \textcolor[HTML]{C49A00}{polyfw}  &  $0.0391_{(0.0033)}$ &  $0.0391_{(0.0033)}$ &  $0.0391_{(0.0033)}$ &  $0.0391_{(0.0033)}$ &  $0.0391_{(0.0033)}$ &  $0.0391_{(0.0033)}$ \\
    \textcolor[HTML]{C49A00}{vrfw}    &  $0.0391_{(0.0033)}$ &  $0.0391_{(0.0033)}$ &  $0.0391_{(0.0033)}$ &  $0.0391_{(0.0033)}$ &  $0.0391_{(0.0033)}$ &  $0.0391_{(0.0033)}$ \\
    \textcolor[HTML]{C49A00}{htfw}    &  $0.0391_{(0.0033)}$ &  $0.0391_{(0.0033)}$ &  $0.0391_{(0.0033)}$ &  $0.0391_{(0.0033)}$ &  $0.0391_{(0.0033)}$ &  $0.0391_{(0.0033)}$ \\
    \textcolor[HTML]{C49A00}{htpl}    &  $0.0391_{(0.0033)}$ &  $0.0391_{(0.0033)}$ &  $0.0391_{(0.0033)}$ &  $0.0391_{(0.0033)}$ &  $0.0391_{(0.0033)}$ &  $0.0385_{(0.0033)}$ \\
    \hline[dashed]
    \textcolor[HTML]{00B6EB}{dpight}  &   $0.0390_{(0.0033)}$ &   $0.0390_{(0.0033)}$ &   $0.0390_{(0.0033)}$ &  $0.0391_{(0.0033)}$ &  $0.0391_{(0.0033)}$ &  $0.0391_{(0.0033)}$ \\
    \textcolor[HTML]{00B6EB}{dpslkt}  &  $0.0391_{(0.0033)}$ &    $0.0400_{(0.0035)}$ &  $0.0394_{(0.0034)}$ &  $0.0391_{(0.0033)}$ &  $0.0391_{(0.0033)}$ &  $0.0391_{(0.0033)}$ \\
    \textcolor[HTML]{00B6EB}{htsl}    &  $0.0384_{(0.0033)}$ &  $0.0388_{(0.0033)}$ &  $0.0388_{(0.0033)}$ &  $0.0389_{(0.0033)}$ &   $0.0390_{(0.0033)}$ &  $0.0391_{(0.0033)}$ \\
    \textcolor[HTML]{00B6EB}{htso}    &  $\mathbf{0.0308}_{(0.0031)}$ &  $\mathbf{0.0247}_{(0.0043)}$ &  $\mathbf{0.0229}_{(0.0044)}$ &   $\mathbf{0.0246}_{(0.0040)}$ &  $\mathbf{0.0282}_{(0.0034)}$ &  $0.0392_{(0.0034)}$ \\
    \hline[dashed]
    \textcolor[HTML]{A58AFF}{gcdgsq}  &  $0.0909_{(0.0132)}$ &  $0.0658_{(0.0064)}$ &  $0.0657_{(0.0064)}$ &  $0.0656_{(0.0064)}$ &  $0.0656_{(0.0064)}$ &  $0.0566_{(0.0058)}$ \\
    \textcolor[HTML]{A58AFF}{gcdgsr}  &  $0.0656_{(0.0064)}$ &  $0.0656_{(0.0064)}$ &  $0.0656_{(0.0064)}$ &  $0.0656_{(0.0064)}$ &  $0.0656_{(0.0064)}$ &  $0.0656_{(0.0064)}$ \\
    \textcolor[HTML]{A58AFF}{gcdgss}  &  $0.6121_{(0.1639)}$ &  $0.0834_{(0.0085)}$ &   $0.0690_{(0.0061)}$ &   $0.0660_{(0.0064)}$ &  $0.0656_{(0.0064)}$ &  $0.0656_{(0.0064)}$ \\
    \hline[dashed]
    \textcolor[HTML]{FB61D7}{nm}      &  $0.0391_{(0.0033)}$ &  $0.0391_{(0.0033)}$ &  $0.0391_{(0.0033)}$ &  $0.0391_{(0.0033)}$ &  $0.0391_{(0.0033)}$ &  $0.0391_{(0.0033)}$ \\
    \hline[dashed]
    sgd      &  $4.0000_{(0.0000)}$ &  $0.0391_{(0.0033)}$ &  $0.0391_{(0.0033)}$ &  $0.0391_{(0.0033)}$ &  $0.0391_{(0.0033)}$ &  $0.0391_{(0.0033)}$ \\
    \bottomrule
    \end{tblr}
    \end{sc}
    \end{small}
\end{table*}

\subsection{Logistic Regression}

Table V, Table VI, and Table VII demonstrate the mean of 20 trials of accuracies for the test portions of the Heart, DBworld, and RCV1 datasets, respectively. Subscripts are provided denoting two times the standard error of the mean. Note that ADMM, ADMMHalf, and ProjERM are not included in Table VII since they did not converge within 48 hours. Additionally, the ECOS solver failed to converge in the TS procedure in Table VI. These three tables are presented in the appendix due to the page limit of the pain body of the paper. Note that in the appendix, Figure 5, Figure 6, and Figure 7 provide boxplots of the accuracies for the trials on each dataset. 

From these three tables, it is clear that GCD, or greedy coordinate descent, performs well on all values of $\epsilon$. As discussed in the previous subsection, GCD methods do not rely on the Lipschitz constant for optimization of the entire weight vector but instead use different Lipschitz constants for each coefficient. This allows GCD to add less noise than other optimizers. 

\noindent\fbox{\parbox{\linewidth}{
\hspace{0.30em} Thus we conclude this section similar to that of linear regression: future research on private logistic regression models should incorporate the scale of each feature separately when adding private noise. This allows for improved empirical utility, in which features with different scales can be noised effectively. 
}}

\section{Trends in Results} \label{sec:trends}

Having completed the implementation and empirical evaluation of each surveyed method, we note three high-level themes in the research that may form valuable directions for future work. Each of these is significant in that they inhibit practitioner adoption of DP methods by imposing unexpected costs of behaviors compared to non-private methods. 

First is the extreme computational cost of using DP linear models compared to their original non-private counterparts, taking orders of magnitude more time to train. This limits real-world utility where datasets can not be subsampled for convenience. A related issue is that reguarlizaition penalties do not have the same (mostly) intuitive behavior in a DP model, and there appears to be no clear guidance on how and when to apply regularizers to a DP linear model. Finally, we find that accuracy does not necessarily decrease with stricter privacy parameters. We discuss these issues in the following subsections. 

\subsection{Computational Inefficiency}

We have evaluated six values of $\epsilon$ for $\approx$ 16 algorithms for each table. Filling each cell required 12-36 hours of computing time for datasets that are considered small by modern machine learning standards. Indeed, training a linear model via scikit-learn or other APIs takes minutes. This extreme cost disparity is perhaps the biggest hindrance to the adoption of DP methods in practice, as a practitioner has no reasonable expectation that they could train a model on a real-world larger dataset. 

As it exists today, only one work has attempted to make the FW algorithm scalable to sparse datasets ~\cite{raff2023scaling}. While they were able to obtain significant speedups of up to four orders of magnitude, the FW algorithm never performed best in any of the $\epsilon \leq 5$ experiments. It remains to be determined how many DP algorithms for linear models can be made computationally tractable while balancing their efficacy. A key issue in this task is the nature of DP itself; addining noise at each step of the process is a computationally demanding task in-and-of itself that tends to remove all sparsity that one would exploit in high dimensional problems. 

\subsection{Unclear Impact of Regularization}

In the linear model case, the $L_1$ penalty is particularly desirable because it is robust to spurious uninformative features and produces sparse solutions ~\cite{Ng_2004}. For both $L_1$ and $L_2$ models, a regularization path is a common desiderata where the penalty $\lambda$ is varied from small-to-large, and a smooth relationship between coefficients (and accuracy) with the value of $\lambda$ is used to inform the model selection and interrogate the data~\cite{Friedman_Hastie_Tibshirani_2010}. This is often achieved via a ``warm start'' where the solution of one value of $\lambda$ is used as the starting solution for an adjacent value of $\lambda' \neq \lambda$ ~\cite{Tsai_Lin_Lin_2014}. 

When training differentially private linear models, we do not see these common benefits occur. In all cases, even when using a $L_1$ penalty on the weight vector, the solution is completely dense. This simultaneously removes the benefit of sparse solutions (larger models) and mitigates the benefit of robustness in the face of irrelevant features (because all features are always used). Warm starting the solution via a previous value of $\lambda$ is also non-viable as it would require expending additional privacy budget. 

The nature of how to regularize DP linear models and their ultimate benefit is an open problem. One possible hypothesis is that the noise added by DP itself acts as a regularize. This is supported by prior results in deep learning on the utility of noise ~\cite{10.5555/3295222.3295264}. Other works in DP learning with a backbone network trained on non-private data also support an implicit regularization effect in obtaining good results without any additional regularization ~\cite{tobaben2023efficacy}, though the use of a backbone confounds drawing a conclusion.

\subsection{Accuracy Asperity with Privacy}

It is natural that as $\epsilon$ decreases, the accuracy should also decrease, as the limit of $\epsilon \rightarrow 0$ implies a constant prediction. However, this is only a weakly observable trend in our experiments despite running each trial 20 times and averaging the result. This has also been observed by \cite{10.1145/3576915.3623081, khanna2022privacy}, and appears to be a broader issue with DP learning. To some degree the issue is unavoidable because noise is added to the model, and so noise necessarily increases in the output. However, it remains to be determined if the amount of asperity in results for a given change in $\epsilon$ is ``optimal'' and how much this could be reduced. 

One possible issue is the intersection of the optimization algorithm used, and that of the noise added. Running a classical optimizer until convergence to a global minima (in the case of strongly convex losses) is possible regardless of the convergence rate of the algorithm being tested. This is not the case for DP, as each access to the data must be accounted for in the total privacy budget. This creates an interplay between target $\epsilon$ and convergence where it may be best to perform fewer optimization iterations to achieve a lower total $\epsilon$, but thus result in an irregular effective convergence rate. Another possibility is implied by the results of ~\cite{10.1145/3605764.3623904} who observe that there is a finite practical range of $\epsilon$ values that have a meaningful impact on privacy. Values of $\epsilon$ below or above this range add/reduce the noise without changing the identifiability of the data and thus could lead to lower $\epsilon$ values obtaining better accuracy by chance because chance was the only factor in relative ranking. 

\section{Conclusion}

In this paper, we provide the first unified review of optimization methods for high dimensional DP linear models. In doing so, we give an overview of the strengths and weaknesses of many methods, highlighting the different approaches taken in prior literature. Next, we conduct empirical experiments for a systematic comparison of the optimization methods, which has not been done in works prior to this. Finally, we release our code for easier future use and better analysis of future algorithms. 

Our empirical experiments highlight surprising and previously unobserved trends of optimization for high-dimensional DP linear models. Specifically, we find that methods which are able to take into account the scale of each feature perform better than those which rely on the Lipschitz constant of the loss function. Indeed, even when the Lipschitz constant of the loss function is bounded, heavy-tailed or coordinate-based optimization techniques can perform better since they are more robust and add less noise. 

We believe that this paper and the surprising result highlighted above can influence future research on differentially private optimization, which is an active research field. Further study on heavy-tailed or coordinate-wise optimization can improve performance on the tasks listed in this review, and may even translate to more complicated models such as neural networks. 

\bibliographystyle{IEEEtran}
\bibliography{IEEEabrv, bibliography}

\clearpage
\appendix
First, we provide tabular results for the logistic regression datasets which did not fit into the main body of the paper:

\begin{table*}[!h]
    \caption{\textbf{Heart}: Accuracy}
    \label{tab:my_label}
    \centering
    \begin{small}
    \begin{sc}
    \SetTblrInner{rowsep=0pt}
    \begin{tblr}{lcccccc}
    \toprule
    $\epsilon$ &                      0.1   &                   0.5   &                   1.0   &                2.0   &                5.0   &                Nonprivate \\
    \midrule
    \textcolor[HTML]{F8766D}{ts}       &  $0.4639_{(0.0823)}$ &  $0.4713_{(0.0829)}$ &  $0.6509_{(0.0475)}$ &  $0.6722_{(0.0347)}$ &  $0.7333_{(0.0258)}$ &  $0.8009_{(0.0237)}$ \\
    \hline[dashed]
    \textcolor[HTML]{C49A00}{fw}       &  $0.5481_{(0.0451)}$ &   $0.5778_{(0.0530)}$ &   $0.5806_{(0.0540)}$ &   $0.6130_{(0.0262)}$ &  $0.6509_{(0.0304)}$ &  $0.7602_{(0.0253)}$ \\
    \textcolor[HTML]{C49A00}{polyfw}   &  $0.5759_{(0.0577)}$ &  $0.5759_{(0.0577)}$ &  $0.5759_{(0.0577)}$ &  $0.5759_{(0.0577)}$ &  $0.5759_{(0.0577)}$ &   $0.5620_{(0.0658)}$ \\
    \textcolor[HTML]{C49A00}{vrfw}     &  $0.3815_{(0.0264)}$ &  $0.3815_{(0.0264)}$ &  $0.3815_{(0.0264)}$ &  $0.3815_{(0.0264)}$ &  $0.3815_{(0.0264)}$ &   $0.5630_{(0.0437)}$ \\
    \textcolor[HTML]{C49A00}{htfw}     &  $0.6602_{(0.0544)}$ &  $0.6648_{(0.0519)}$ &  $0.6602_{(0.0544)}$ &  $0.6602_{(0.0544)}$ &    $0.6630_{(0.0530)}$ &     $0.8000_{(0.0295)}$ \\
    \hline[dashed]
    \textcolor[HTML]{53B400}{projerm}  &  $0.4537_{(0.0724)}$ &  $0.4537_{(0.0724)}$ &  $0.4481_{(0.0741)}$ &  $0.4481_{(0.0741)}$ &  $0.4509_{(0.0739)}$ &  $0.7741_{(0.0229)}$ \\
    \hline[dashed]
    \textcolor[HTML]{00C094}{admm}     &  $0.6519_{(0.0308)}$ &  $0.6815_{(0.0453)}$ &  $0.6843_{(0.0507)}$ &  $0.6815_{(0.0526)}$ &  $0.6861_{(0.0357)}$ &   $0.6380_{(0.0282)}$ \\
    \textcolor[HTML]{00C094}{admmhalf} &  $\mathbf{0.7861}_{(0.0232)}$ &  $0.7963_{(0.0237)}$ &  $0.7972_{(0.0237)}$ &  $0.7944_{(0.0245)}$ &   $\mathbf{0.8019}_{(0.0240)}$ &  $0.7602_{(0.0205)}$ \\
    \hline[dashed]
    \textcolor[HTML]{00B6EB}{dpight}   &  $0.5787_{(0.0508)}$ &  $0.5944_{(0.0423)}$ &   $0.6880_{(0.0428)}$ &   $0.7380_{(0.0493)}$ &  $0.7991_{(0.0228)}$ &  $0.8278_{(0.0223)}$ \\
    \textcolor[HTML]{00B6EB}{dpslkt}   &  $0.5981_{(0.0396)}$ &   $0.5843_{(0.0390)}$ &  $0.5981_{(0.0396)}$ &  $0.5981_{(0.0396)}$ &  $0.5935_{(0.0384)}$ &  $0.7611_{(0.0245)}$ \\
    \textcolor[HTML]{00B6EB}{htso}     &  $0.6722_{(0.0294)}$ &  $0.6731_{(0.0301)}$ &  $0.6731_{(0.0301)}$ &  $0.6759_{(0.0292)}$ &  $0.6769_{(0.0313)}$ &   $\mathbf{0.8287}_{(0.0240)}$ \\
    \hline[dashed]
    \textcolor[HTML]{A58AFF}{gcdgsq}   &  $0.7852_{(0.0235)}$ &  $\mathbf{0.8306}_{(0.0261)}$ &  $\mathbf{0.8111}_{(0.0229)}$ &  $\mathbf{0.8111}_{(0.0229)}$ &  $0.7954_{(0.0216)}$ &  $0.7722_{(0.0325)}$ \\
    \textcolor[HTML]{A58AFF}{gcdgsr}   &  $0.7778_{(0.0206)}$ &  $0.7778_{(0.0206)}$ &  $0.7778_{(0.0206)}$ &  $0.7778_{(0.0206)}$ &  $0.7778_{(0.0206)}$ &  $0.7778_{(0.0201)}$ \\
    \textcolor[HTML]{A58AFF}{gcdgss}   &  $0.7574_{(0.0203)}$ &  $0.7574_{(0.0203)}$ &  $0.7574_{(0.0203)}$ &  $0.7574_{(0.0203)}$ &  $0.7574_{(0.0203)}$ &  $0.7574_{(0.0203)}$ \\
    \hline[dashed]
    \textcolor[HTML]{FB61D7}{nm}       &   $0.5324_{(0.0540)}$ &     $0.3000_{(0.0424)}$ &  $0.3704_{(0.0294)}$ &  $0.3111_{(0.0362)}$ &  $0.4259_{(0.0415)}$ &  $0.3778_{(0.0451)}$ \\
    \hline[dashed]
    sgd      &  $0.6269_{(0.0464)}$ &  $0.6269_{(0.0464)}$ &  $0.6269_{(0.0464)}$ &  $0.6269_{(0.0464)}$ &  $0.6269_{(0.0464)}$ &  $0.6713_{(0.0306)}$ \\
    \bottomrule
    \end{tblr}
    \end{sc}
    \end{small}
\end{table*}

\begin{table*}[!h]
    \caption{\textbf{DBworld}: Accuracy}
    \label{tab:my_label}
    \centering
    \begin{small}
    \begin{sc}
    \SetTblrInner{rowsep=0pt}
    \begin{tblr}{lcccccc}
    \toprule
    $\epsilon$ &                      0.1   &                   0.5   &                   1.0   &                2.0   &                5.0   &                Nonprivate \\
    \midrule
    \textcolor[HTML]{C49A00}{fw}       &  $0.4769_{(0.0637)}$ &  $0.5077_{(0.0504)}$ &  $0.4769_{(0.0637)}$ &    $0.4808_{(0.06)}$ &  $0.4769_{(0.0637)}$ &  $\mathbf{0.7038}_{(0.0477)}$ \\
    \textcolor[HTML]{C49A00}{polyfw}   &  $0.4692_{(0.0557)}$ &  $0.4692_{(0.0557)}$ &  $0.4692_{(0.0557)}$ &   $0.5077_{(0.054)}$ &  $0.4692_{(0.0557)}$ &   $0.4769_{(0.038)}$ \\
    \textcolor[HTML]{C49A00}{vrfw}     &  $0.6038_{(0.0561)}$ &  $0.6038_{(0.0561)}$ &  $0.6038_{(0.0561)}$ &  $0.6038_{(0.0561)}$ &  $0.6038_{(0.0561)}$ &  $0.5077_{(0.0573)}$ \\
    \textcolor[HTML]{C49A00}{htfw}     &  $0.4731_{(0.0464)}$ &  $0.4731_{(0.0464)}$ &  $0.4731_{(0.0464)}$ &  $0.4731_{(0.0464)}$ &  $0.4731_{(0.0464)}$ &  $0.5154_{(0.0462)}$ \\
    \hline[dashed]
    \textcolor[HTML]{53B400}{projerm}  &    $0.5500_{(0.0614)}$ &  $0.5808_{(0.0655)}$ &  $0.5808_{(0.0655)}$ &  $0.5808_{(0.0655)}$ &  $0.5808_{(0.0655)}$ &  $0.6423_{(0.0572)}$ \\
    \hline[dashed]
    \textcolor[HTML]{00C094}{admm}     &  $0.4808_{(0.0669)}$ &  $0.4808_{(0.0669)}$ &  $0.4808_{(0.0669)}$ &  $0.5115_{(0.0653)}$ &  $0.4808_{(0.0669)}$ &  $0.5615_{(0.0724)}$ \\
    \textcolor[HTML]{00C094}{admmhalf} &  $0.5462_{(0.0472)}$ &    $0.5500_{(0.0538)}$ &    $0.5500_{(0.0464)}$ &  $0.5731_{(0.0646)}$ &  $0.4885_{(0.0358)}$ &  $0.5615_{(0.0724)}$ \\
    \hline[dashed]
    \textcolor[HTML]{00B6EB}{dpight}   &  $0.5308_{(0.0546)}$ &   $0.4731_{(0.0450)}$ &   $0.5885_{(0.0490)}$ &  $0.5077_{(0.0573)}$ &   $0.4808_{(0.043)}$ &   $0.5462_{(0.074)}$ \\
    \textcolor[HTML]{00B6EB}{dpslkt}   &  $0.4654_{(0.0467)}$ &  $0.4654_{(0.0467)}$ &  $0.4654_{(0.0467)}$ &  $0.4654_{(0.0467)}$ &   $0.5462_{(0.059)}$ &  $0.4962_{(0.0541)}$ \\
    \textcolor[HTML]{00B6EB}{htso}     &  $0.6038_{(0.0561)}$ &  $0.6038_{(0.0561)}$ &  $0.6115_{(0.0683)}$ &  $0.6192_{(0.0626)}$ &  $0.6192_{(0.0626)}$ &  $0.5154_{(0.0592)}$ \\
    \hline[dashed]
    \textcolor[HTML]{A58AFF}{gcdgsq}   &  $\mathbf{0.6885}_{(0.0378)}$ &  $\mathbf{0.7154}_{(0.0448)}$ &  $\mathbf{0.7154}_{(0.0448)}$ &   $\mathbf{0.6769}_{(0.077)}$ &  $\mathbf{0.7154}_{(0.0652)}$ &  $0.6885_{(0.0505)}$ \\
    \textcolor[HTML]{A58AFF}{gcdgsr}   &  $0.5615_{(0.0536)}$ &  $0.5615_{(0.0536)}$ &  $0.5615_{(0.0536)}$ &  $0.5615_{(0.0536)}$ &  $0.5615_{(0.0536)}$ &  $0.5615_{(0.0536)}$ \\
    \textcolor[HTML]{A58AFF}{gcdgss}   &  $0.6577_{(0.0736)}$ &  $0.6577_{(0.0736)}$ &  $0.6577_{(0.0736)}$ &  $0.6577_{(0.0736)}$ &  $0.6577_{(0.0736)}$ &     $0.7_{(0.0431)}$ \\
    \hline[dashed]
    \textcolor[HTML]{FB61D7}{nm}       &  $0.5192_{(0.0485)}$ &  $0.5192_{(0.0485)}$ &  $0.5654_{(0.0743)}$ &   $0.4692_{(0.062)}$ &  $0.4154_{(0.0846)}$ &  $0.4808_{(0.0545)}$ \\
    \hline[dashed]
    sgd      &  $0.4154_{(0.0504)}$ &  $0.4154_{(0.0504)}$ &  $0.4154_{(0.0504)}$ &  $0.4154_{(0.0504)}$ &  $0.4154_{(0.0504)}$ &  $0.5154_{(0.0536)}$ \\
    \bottomrule
    \end{tblr}
    \end{sc}
    \end{small}
\end{table*}

\begin{table*}
    \caption{\textbf{RCV1}: Accuracy}
    \label{tab:rcv1}
    \centering
    \begin{small}
    \begin{sc}
    \SetTblrInner{rowsep=0pt}
    \begin{tblr}{lcccccc}
    \toprule
    $\epsilon$ &                      0.1   &                   0.5   &                   1.0   &                2.0   &                5.0   &                Nonprivate \\
    \midrule
    \textcolor[HTML]{F8766D}{ts}       &  $0.5095_{(0.0578)}$ &    $0.4550_{(0.0540)}$ &    $0.4550_{(0.0540)}$ &  $0.4575_{(0.0541)}$ &   $0.4585_{(0.0540)}$ &   $0.4670_{(0.0531)}$ \\
    \hline[dashed]
    \textcolor[HTML]{C49A00}{fw}       &  $0.5175_{(0.0217)}$ &  $0.4995_{(0.0204)}$ &   $0.4950_{(0.0176)}$ &   $0.4930_{(0.0167)}$ &  $0.5175_{(0.0217)}$ &   $\mathbf{0.6440}_{(0.0323)}$ \\
    \textcolor[HTML]{C49A00}{polyfw}   &  $0.5055_{(0.0183)}$ &  $0.5015_{(0.0218)}$ &  $0.5055_{(0.0183)}$ &   $0.5030_{(0.0202)}$ &  $0.5055_{(0.0183)}$ &   $0.4895_{(0.0230)}$ \\
    \textcolor[HTML]{C49A00}{vrfw}     &    $0.5170_{(0.0190)}$ &    $0.5170_{(0.0190)}$ &    $0.5170_{(0.0190)}$ &    $0.5170_{(0.0190)}$ &    $0.5170_{(0.0190)}$ &   $0.4930_{(0.0263)}$ \\
    \textcolor[HTML]{C49A00}{htfw}     &  $0.5185_{(0.0168)}$ &  $0.5185_{(0.0168)}$ &  $0.5185_{(0.0168)}$ &  $0.5185_{(0.0164)}$ &  $0.5185_{(0.0168)}$ &  $0.5805_{(0.0555)}$ \\
    \hline[dashed]
    \textcolor[HTML]{00B6EB}{dpight}   &  $0.5255_{(0.0212)}$ &   $0.5140_{(0.0274)}$ &   $0.5140_{(0.0274)}$ &   $0.5140_{(0.0274)}$ &   $0.5140_{(0.0274)}$ &  $0.5315_{(0.0248)}$ \\
    \textcolor[HTML]{00B6EB}{dpslkt}   &    $0.5660_{(0.0170)}$ &   $0.5620_{(0.0167)}$ &    $0.5660_{(0.0170)}$ &  $0.5485_{(0.0295)}$ &   $0.5620_{(0.0167)}$ &  $0.5225_{(0.0208)}$ \\
    \textcolor[HTML]{00B6EB}{htso}     &    $0.5060_{(0.0220)}$ &  $0.5195_{(0.0266)}$ &    $0.5360_{(0.0180)}$ &    $0.5360_{(0.0180)}$ &  $0.5195_{(0.0266)}$ &  $0.5145_{(0.0236)}$ \\
    \hline[dashed]
    \textcolor[HTML]{A58AFF}{gcdgsq}   &   $0.5790_{(0.0222)}$ &  $0.5805_{(0.0222)}$ &  $\mathbf{0.5945}_{(0.0213)}$ &    $\mathbf{0.5940}_{(0.0210)}$ &  $0.5955_{(0.0201)}$ &  $0.6125_{(0.0185)}$ \\
    \textcolor[HTML]{A58AFF}{gcdgsr}   &  $0.5305_{(0.0237)}$ &  $0.5305_{(0.0237)}$ &  $0.5305_{(0.0237)}$ &  $0.5305_{(0.0237)}$ &  $0.5285_{(0.0222)}$ &   $0.5860_{(0.0269)}$ \\
    \textcolor[HTML]{A58AFF}{gcdgss}   &   $\mathbf{0.5920}_{(0.0184)}$ &   $\mathbf{0.5920}_{(0.0184)}$ &  $0.5925_{(0.0185)}$ &   $\mathbf{0.5940}_{(0.0187)}$ &   $\mathbf{0.5970}_{(0.0195)}$ &   $0.6180_{(0.0154)}$ \\
    \hline[dashed]
    \textcolor[HTML]{FB61D7}{nm}       &  $0.4985_{(0.0239)}$ &   $0.5010_{(0.0232)}$ &  $0.4745_{(0.0208)}$ &  $0.5135_{(0.0197)}$ &  $0.4615_{(0.0204)}$ &   $0.5020_{(0.0233)}$ \\
    \hline[dashed]
    sgd      &  $0.0000_{(0.0000)}$ &  $0.5295_{(0.0216)}$ &  $0.5295_{(0.0216)}$ &  $0.5295_{(0.0216)}$ &  $0.5295_{(0.0216)}$ &  $0.4870_{(0.0180)}$ \\
    \bottomrule
    \end{tblr}
    \end{sc}
    \end{small}
\end{table*}

\clearpage

Here we detail the hyperparameters tested for each algorithm and acronyms employed in the paper: 

\begin{table}[!h]
    \caption{Hyperparameters}
    \label{tab:my_label}
    \centering
    \begin{small}
    \begin{sc}
    \SetTblrInner{rowsep=0pt}
    \begin{tblr}{ll}
    \toprule
    \textcolor[HTML]{F8766D}{ts}       &  \makecell[l]{$\text{Sparsity} \in \{1, 2, 5, 10\}$ \\ $\text{Reg} \in \{0.001, 0.005, 0.01, 0.05, 0.1, 0.5 \}$}\\
    \hline[dashed]
    \textcolor[HTML]{C49A00}{fw}       &  \makecell[l]{$\text{Iter} \in \{1, 2, 5, 10, 20, 50, 100\}$} \\
    \textcolor[HTML]{C49A00}{polyfw}   &  None \\
    \textcolor[HTML]{C49A00}{vrfw}     &  None \\
    \textcolor[HTML]{C49A00}{htfw}     &  \makecell[l]{$\text{Iter} \in \{1, 2, 5, 10, 20, 50, 100\}$ \\ $s \in \{ 1, 10, 100 \}$} \\
    \textcolor[HTML]{C49A00}{htpl} & \makecell[l]{$\text{Iter} \in \{1, 2, 5, 10, 20, 50, 100\}$} \\
    \hline[dashed]
    \textcolor[HTML]{53B400}{projerm}  &  \makecell[l]{$\text{Latent} \in \{2, 5, 10, 20\}$} \\
    \hline[dashed]
    \textcolor[HTML]{00C094}{admm}     &  \makecell[l]{$\text{Iter} \in \{1, 2, 5, 10, 20, 50, 100\}$ \\ $\gamma \in \{ 0.001, 0.01, 0.1, 1 \}$ \\ $\text{Reg} \in \{0.001, 0.005, 0.01, 0.05, 0.1, 0.5 \}$}\\
    \textcolor[HTML]{00C094}{admmhalf} &   \makecell[l]{$\text{Iter} \in \{1, 2, 5, 10, 20, 50, 100\}$ \\ $\gamma \in \{ 0.001, 0.01, 0.1, 1 \}$ \\ $\text{Reg} \in \{0.001, 0.005, 0.01, 0.05, 0.1, 0.5 \}$}\\
    \hline[dashed]
    \textcolor[HTML]{00B6EB}{dpight}   &  \makecell[l]{$\text{Sparsity} \in \{1, 2, 5, 10\}$ \\ $\text{LR} \in \{ 0.001, 0.01, 0.1 \}$ \\ $\text{Iter} \in \{1, 2, 5, 10, 20, 50, 100\}$} \\ 
    \textcolor[HTML]{00B6EB}{dpslkt}   &  \makecell[l]{$\text{Sparsity} \in \{1, 2, 5, 10\}$ \\ $\text{LR} \in \{ 0.001, 0.01, 0.1 \}$ \\ $\text{Iter} \in \{1, 2, 5, 10, 20, 50, 100\}$ \\ $\lambda \in \{ 0.001, 0.01, 0.1, 1 \}$} \\
    \textcolor[HTML]{00B6EB}{htsl}     &  \makecell[l]{$\text{Sparsity} \in \{1, 2, 5, 10\}$ \\ $\text{LR} \in \{ 0.001, 0.01, 0.1 \}$ \\ $\text{Iter} \in \{1, 2, 5, 10, 20, 50, 100\}$} \\
    \textcolor[HTML]{00B6EB}{htso}     &  \makecell[l]{$\text{Sparsity} \in \{1, 2, 5, 10\}$ \\ $\text{LR} \in \{ 0.001, 0.01, 0.1 \}$ \\ $\text{Iter} \in \{1, 2, 5, 10, 20, 50, 100\}$ \\ $s \in \{1, 10, 100 \}$} \\
    \hline[dashed]
    \textcolor[HTML]{A58AFF}{gcdgsq}   &  \makecell[l]{$\text{Sparsity} \in \{1, 2, 5, 10\}$ \\ $\text{Iter} \in \{1, 2, 5, 10, 20, 50, 100\}$ \\ $\text{Reg} \in \{0.001, 0.005, 0.01, 0.05, 0.1, 0.5 \}$} \\
    \textcolor[HTML]{A58AFF}{gcdgsr}   &  \makecell[l]{$\text{Sparsity} \in \{1, 2, 5, 10\}$ \\ $\text{Iter} \in \{1, 2, 5, 10, 20, 50, 100\}$ \\ $\text{Reg} \in \{0.001, 0.005, 0.01, 0.05, 0.1, 0.5 \}$} \\
    \textcolor[HTML]{A58AFF}{gcdgss}   &  \makecell[l]{$\text{Sparsity} \in \{1, 2, 5, 10\}$ \\ $\text{Iter} \in \{1, 2, 5, 10, 20, 50, 100\}$ \\ $\text{Reg} \in \{0.001, 0.005, 0.01, 0.05, 0.1, 0.5 \}$} \\
    \hline[dashed]
    \textcolor[HTML]{FB61D7}{nm}       &  None  \\
    \hline[dashed]
    sgd & \makecell[l]{$\text{Batch Size} \in \{32, 64, 128\}$ \\ $\text{LR} \in \{0.001, 0.01, 0.1\}$ \\ $\text{Iter} \in \{1, 2, 5, 10, 20, 50, 100\}$} \\ 
    \bottomrule
    \end{tblr}
    \end{sc}
    \end{small}
\end{table}

\begin{table}[!h]
\caption{Acronyms and Definitions}
\label{tbl:acronyms}
\centering
\begin{tabular}{@{}ll@{}}
\toprule
\multicolumn{1}{c}{Acronym} & \multicolumn{1}{c}{Definition} \\ \midrule
DP                          &   Differential Privacy                             \\
TS                          &     Two-Stage \cite{kifer2012private}                           \\
FW                          &   Frank-Wolfe \cite{talwar2015nearly}                             \\
POLYFW                      &  Variance-Reduced Frank-Wolfe \cite{bassily2021non}                              \\
VRFW                        &  Variance-Reduced Frank-Wolfe \cite{asi2021private}                              \\
HTFW                        &  Heavy-Tailed Frank-Wolfe \cite{hu2022high}                              \\
HTPL                        &  Heavy-Tailed Private Lasso \cite{hu2022high}                              \\
PROJERM                     & Projected Empirical Risk Minimization \cite{kasiviswanathan2016efficient}                               \\
ADMM                        & Alternating Direction Method of Multipliers \cite{wang2020differential} \\
ADMMHALF                    & Alternating Direction Method of Multipliers \cite{wang2020differential}                               \\
DPIGHT                      & DP Iterative Gradient Hard Thresholding \cite{wang2019differentially}                               \\
DPSLKT                      & DP Knowledge Transfer Framework \cite{wang2020knowledge}                               \\
HTSL                        & Heavy-Tailed Sparse Lasso \cite{hu2022high}                               \\
HTSO                        & Heavy-Tailed Sparse Optimization \cite{hu2022high}                               \\
GCDGSQ                      & Greedy Coordinate Descent \cite{mangold2022high}                               \\
GCDGSR                      & Greedy Coordinate Descent \cite{mangold2022high}                               \\
GCDGSS                      & Greedy Coordinate Descent \cite{mangold2022high}                               \\
NM                          & Noise Multiplier \cite{asi2021private}                               \\
GCD                         & Greedy Coordinate Descent \cite{mangold2022high}                               \\
CVXPY                       & Convex Programming Python Package \cite{diamond2016cvxpy}                              \\
SCS                         & Splitting Conic Solver \cite{agrawal2018rewriting}                               \\
ECOS                        & Embedded Conic Solver \cite{domahidi2013ecos}                               \\
ERM                         & Empirical Risk Minimization                               \\
CPU                         & Central Processing Unit                               \\
GHz                         & Gigahertz                               \\
PCA                         & Principal Components Analysis                               \\ \bottomrule
\end{tabular}
\end{table}

\clearpage
Next, we provide boxplots summarizing the information provided in the results tables: 

\begin{figure*}
    \centering
    \includegraphics[width=0.85\textwidth]{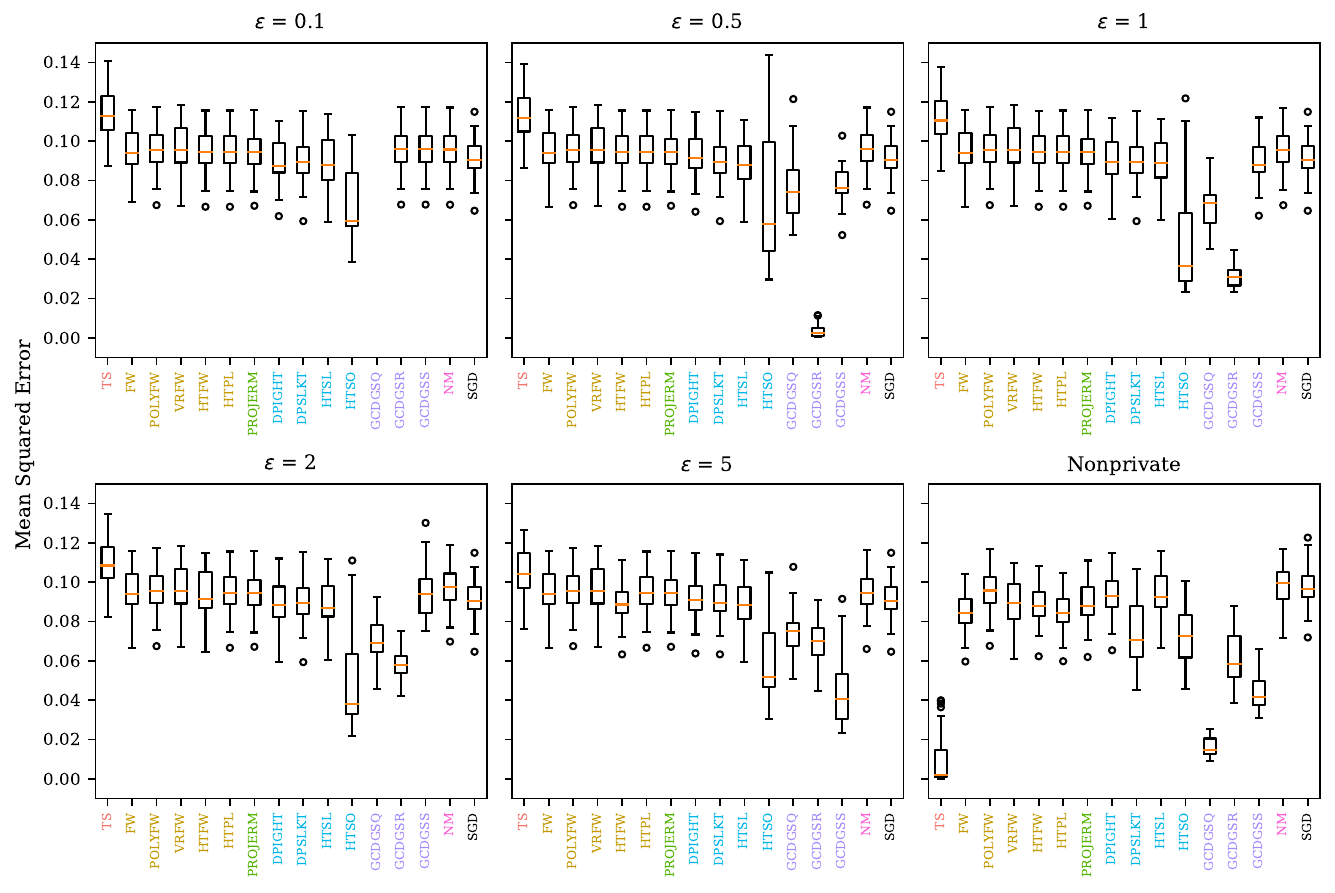}
    \caption{\textbf{Bodyfat}: Mean Squared Error}
\end{figure*}

\begin{figure*}
    \centering
    \includegraphics[width=0.85\textwidth]{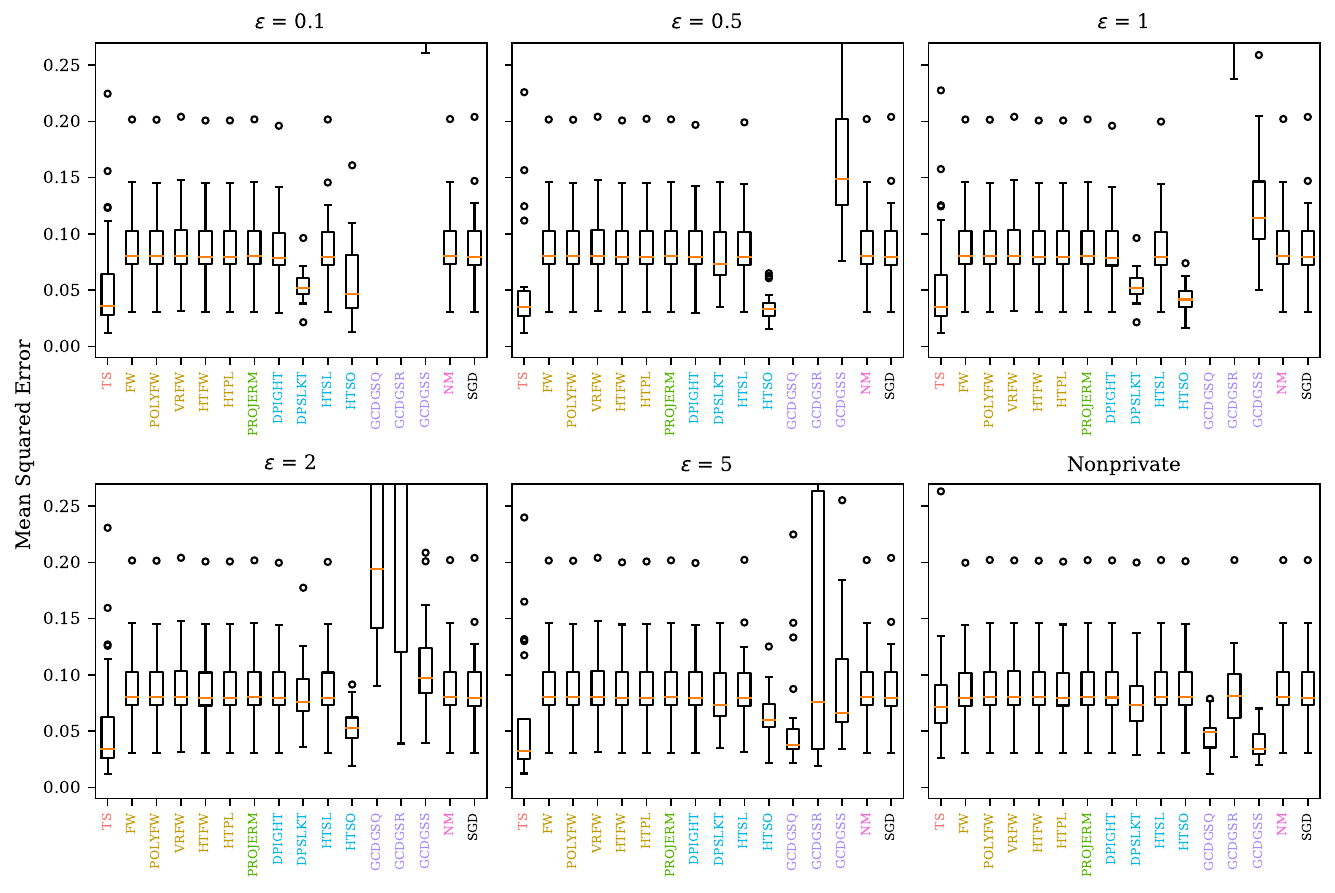}
    \caption{\textbf{PAH}: Mean Squared Error}
\end{figure*}

\begin{figure*}
    \centering
    \includegraphics[width=0.85\textwidth]{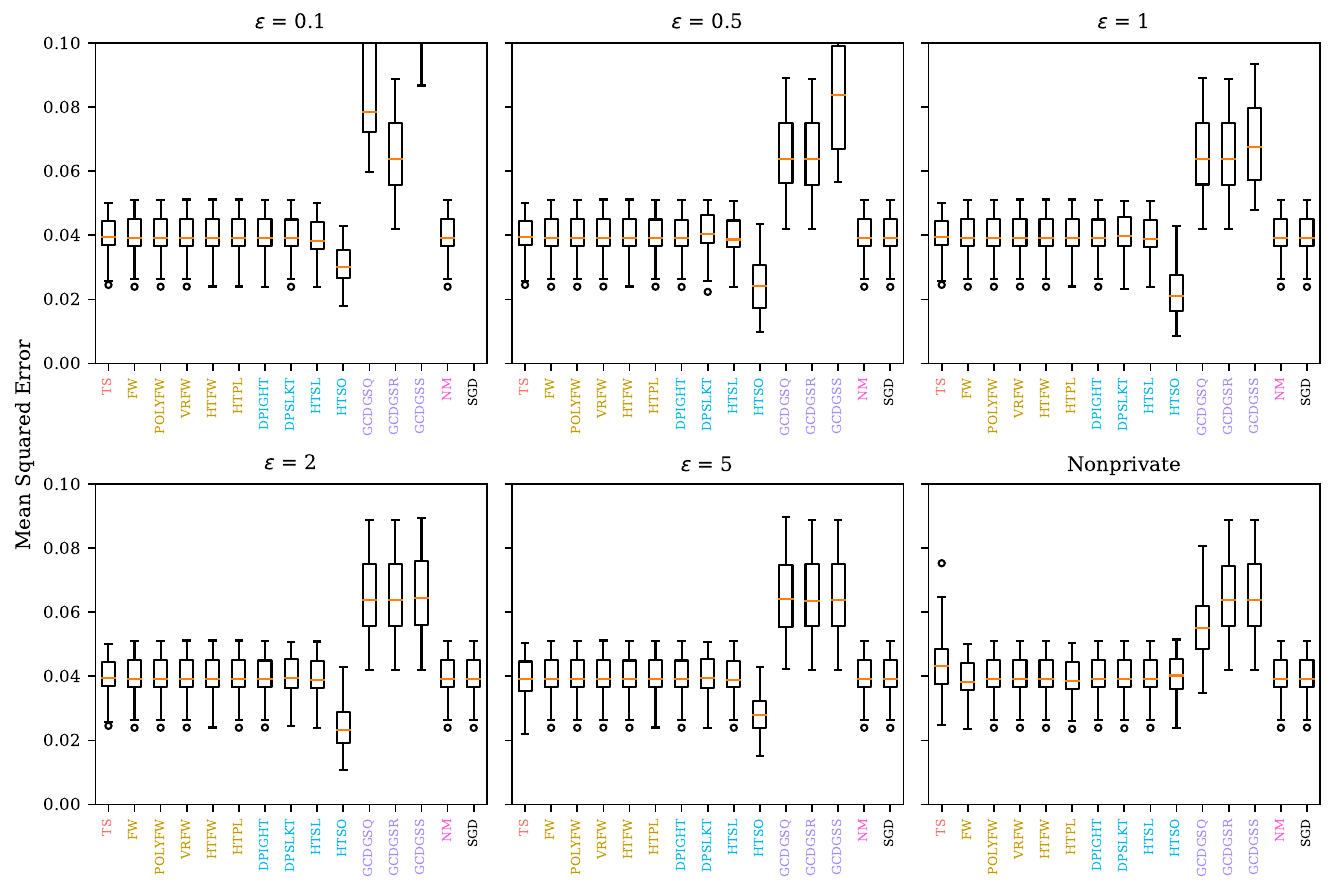}
    \caption{\textbf{E2006}: Mean Squared Error}
\end{figure*}

\begin{figure*}
    \centering
    \includegraphics[width=0.85\textwidth]{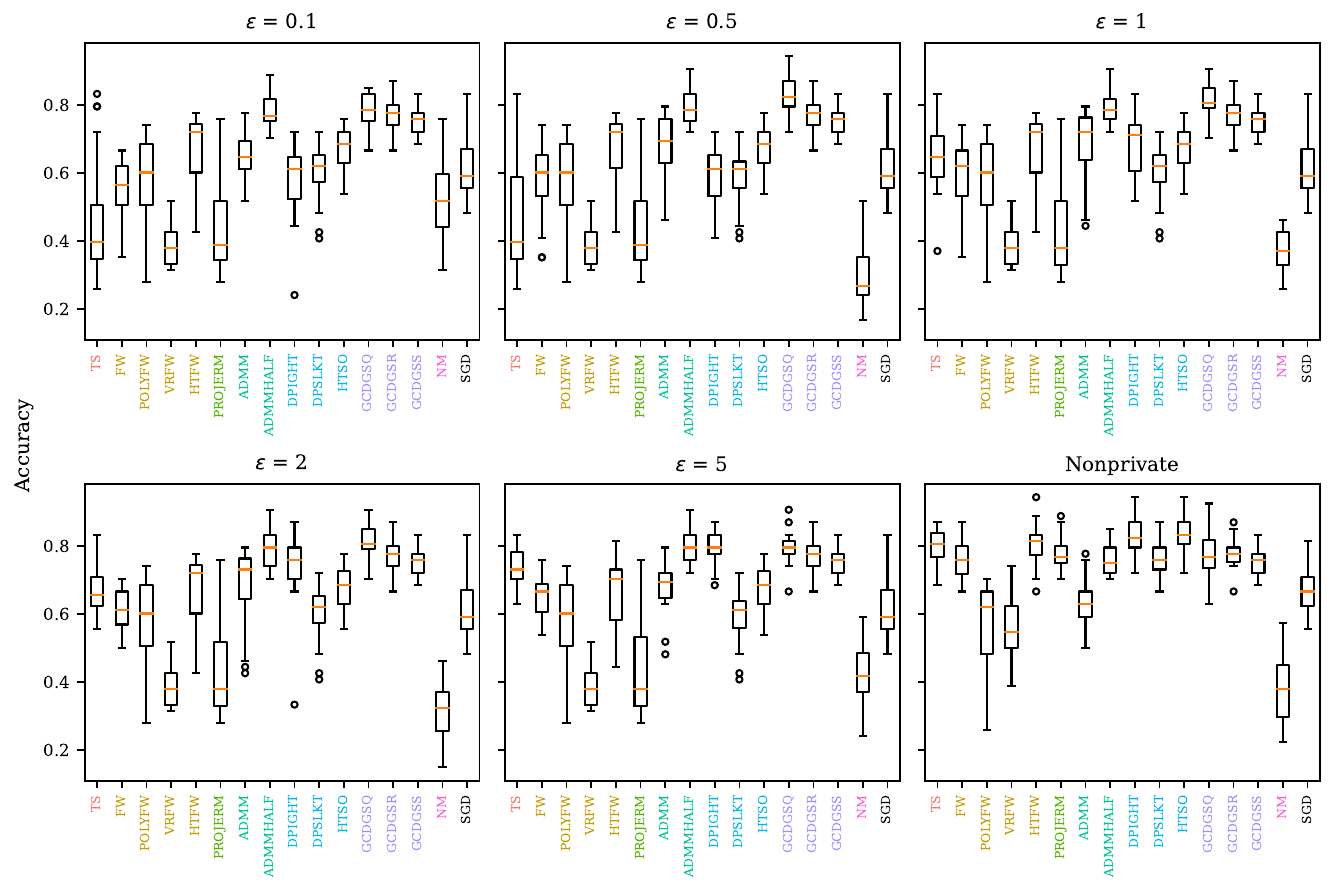}
    \caption{\textbf{Heart}: Accuracy}
\end{figure*}

\begin{figure*}
    \centering
    \includegraphics[width=0.85\textwidth]{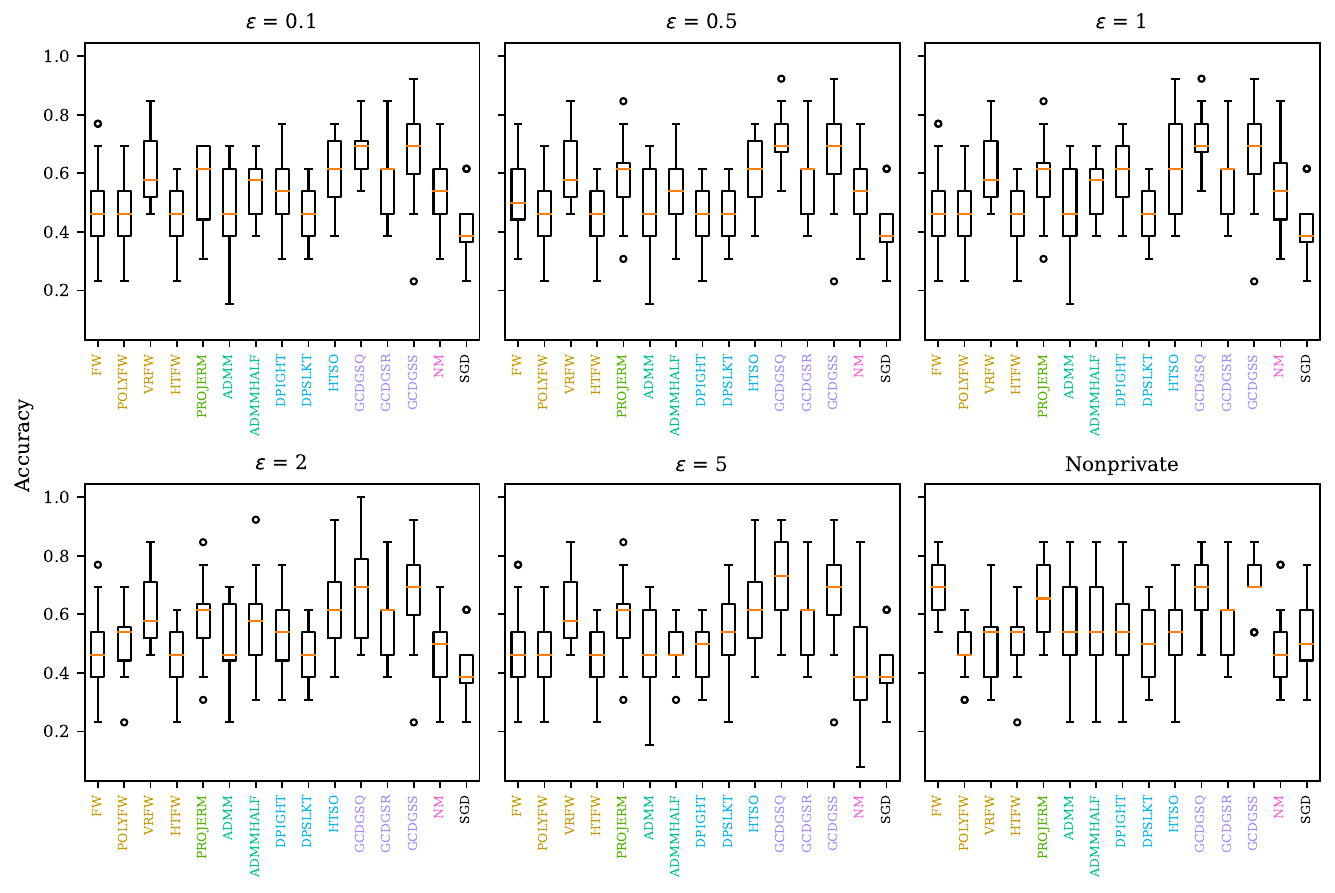}
    \caption{\textbf{DBworld}: Accuracy}
\end{figure*}

\begin{figure*}
    \centering
    \includegraphics[width=0.85\textwidth]{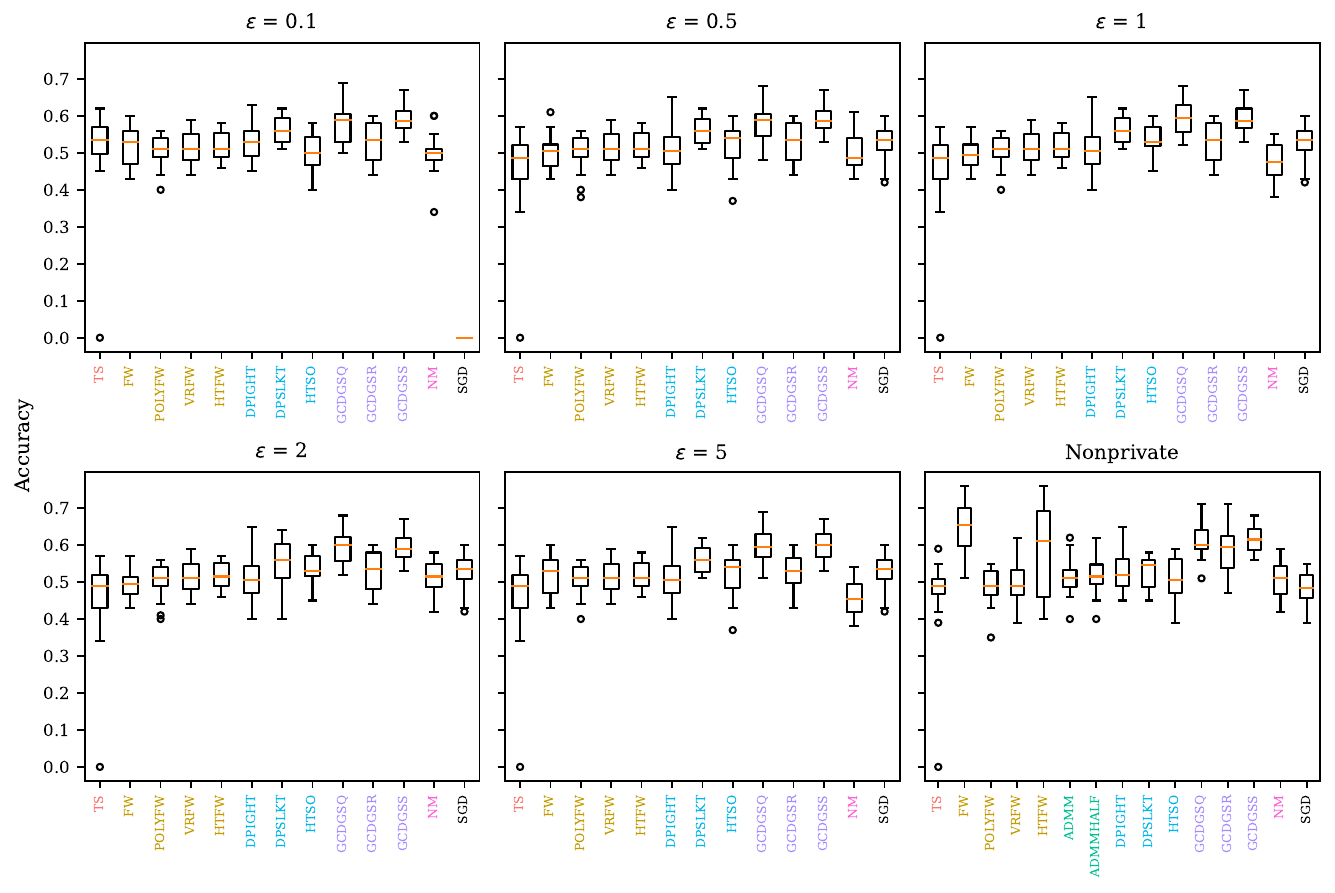}
    \caption{\textbf{RCV1}: Accuracy}
\end{figure*}

\end{document}